%% file: main.tex
\documentclass[10pt,twocolumn,letterpaper]{article}
\input{sec/header}

\title{FedSOL: Stabilized Orthogonal Learning with Proximal \\Restrictions in Federated Learning}

\author {
    Gihun Lee\textsuperscript{\rm 1},
    Minchan Jeong\textsuperscript{\rm 1},
    Sangmook Kim\textsuperscript{\rm 2},
    Jaehoon Oh\textsuperscript{\rm 3},
    Se-Young Yun\textsuperscript{\rm 1}\\
    \textsuperscript{\rm 1}Graduate School of AI \, KAIST \textsuperscript{\rm 3}Samsung Advanced Institute of Technology\\
\textsuperscript{\rm 2}Department of Electrical and Computer Engineering, UBC\\
}

\begin{document}

\maketitle
%
%

\input{sec/00-abstract}
\input{sec/01-intro}
\input{sec/02-motiv}
\input{sec/03-method}
\input{sec/04-experiment}
\input{sec/05-analysis}
\input{sec/06-related-work}
\input{sec/07-conclusion}

\clearpage
\section*{Acknowledgement}
This work was supported by Institute of Information \& communications Technology Planning \& Evaluation (IITP) grant funded by Korea government (MSIT) [No. 2021-0-00907, Development of Adaptive and Lightweight Edge-Collaborative Analysis Technology for Enabling Proactively Immediate Response and Rapid Learning, 90\%] and [No. 2019-0-00075, Artificial Intelligence Graduate School Program (KAIST), 10\%].
{
\small
\bibliographystyle{cvpr}
\bibliography{main}
}

\clearpage
\input{sec/appendix}
\end{document}

%% file: sec/header.tex
\usepackage{cvpr}              

\usepackage[dvipsnames]{xcolor}

\usepackage{amsmath,amssymb,amsfonts}
\usepackage{textcomp}

\usepackage{graphicx}
\usepackage{mwe}
\usepackage{wrapfig}

\usepackage{algorithmic}
\usepackage{algorithm}
\usepackage{tikz}
\usetikzlibrary{fit,calc}

\usepackage{tabularx}
\usepackage{tabulary}
\usepackage{booktabs}
\usepackage{multirow}
\usepackage{arydshln}

\newtheorem{theorem}{Theorem}
\newtheorem{assumption}{Assumption}
\newtheorem{proposition}{Proposition}

\usepackage{lipsum}
\usepackage{url}
\usepackage{blindtext}
\newenvironment{proof}{\paragraph{Proof:}}{\hfill$\square$}

\newcommand{\bw}{\boldsymbol{w}}

\newcommand{\bwk}{\boldsymbol{w}_k}

\newcommand{\bwg}{\boldsymbol{w}_g}
\makeatletter
\def\zzunderbrace#1#2{\mathop{\vtop{\m@th\ialign{##\crcr
   $\hfil\displaystyle{#1}\hfil$\crcr
   \noalign{\kern3\p@\nointerlineskip}%
   \zzupbracefill{#2}\crcr\noalign{\kern3\p@}}}}\limits}

\def\zzupbracefill#1{$\m@th \setbox\z@\hbox{$\braceld$}%
  \bracelu\leaders\vrule \@height\ht\z@ \@depth\z@\hfill
  \,\lower .1em\hbox{\scriptsize#1}\,%
  \leaders\vrule \@height\ht\z@ \@depth\z@\hfill\braceru$}
\makeatletter

\newcommand{\Autoref}[1]{%
  \begingroup%
  \def\algorithmautorefname{Algorithm}%
  \def\chapterautorefname{Chapter}%
  \def\sectionautorefname{Section}%
  \def\subsectionautorefname{Section}%
  \autoref{#1}%
  \endgroup%
}

\definecolor{cvprblue}{rgb}{0.21,0.49,0.74}
\usepackage[pagebackref,breaklinks,colorlinks,citecolor=cvprblue]{hyperref}

\def\paperID{1898} 
\def\confName{CVPR}
\def\confYear{2024}

%% file: sec/00-abstract.tex
\begin{abstract}
Federated Learning\,(FL) aggregates locally trained models from individual clients to construct a global model. While FL enables learning a model with data privacy, it often suffers from significant performance degradation when clients have heterogeneous data distributions. This data heterogeneity causes the model to forget the global knowledge acquired from previously sampled clients after being trained on local datasets. Although the introduction of proximal objectives in local updates helps to preserve global knowledge, it can also hinder local learning by interfering with local objectives. To address this problem, we propose a novel method, Federated Stabilized Orthogonal Learning\,(FedSOL), which adopts an orthogonal learning strategy to balance the two conflicting objectives. FedSOL is designed to identify gradients of local objectives that are inherently orthogonal to directions affecting the proximal objective. Specifically, FedSOL targets parameter regions where learning on the local objective is minimally influenced by proximal weight perturbations. Our experiments demonstrate that FedSOL consistently achieves state-of-the-art performance across various scenarios. 
\end{abstract}
\vspace{-5pt}

%% file: sec/01-intro.tex
\section{Introduction}
\vspace{-2pt}

Federated Learning\,(FL) is an emerging distributed learning framework that preserves data privacy while leveraging client data for training\,\,\citep{federated_learning, federated_optimization}. In this approach, individual clients train their local models using their private data, and a server aggregates these models into a global model. FL eliminates the need for direct access to clients' raw data, enabling the use of extensive data collected from various sources such as mobile phones, vehicles, and facilities\,\citep{fl_survey_technologies_applications, federatedML_concepts_applications}. However, FL encounters a notorious challenge known as data heterogeneity\,\citep{On_the_convergence_fedavg_noniid, fl_challenges_methods, advances_open_problems_fl}. As data from different clients often come from diverse underlying distributions, local datasets are not independently and identically distributed (Non-IID). This common issue in real-world scenarios leads to a misalignment between global and local objectives\,\citep{fednova, scaffold}. Consequently, local learning deviates from the global objective, resulting in significantly degraded performance and slower convergence\,\citep{advances_open_problems_fl, fl_with_noniid}.

Recent research suggests that such deviation in FL resembles \textit{Catastrophic Forgetting} in Continual Learning\,(CL)\,\citep{fedntd, fedcurv, forgetting_survey}. In CL, fitting the model to a new task alters parameters critical for previous tasks, thus impairing their performance\,\citep{cl_forgetting1, cl_forgetting2, cl_survey2, sp_dilemma}. Similarly in FL, local learning is prone to be overfitted on local datasets, causing the model to forget global knowledge not represented in the local distributions\,\citep{fedntd, fedreg}. Consequently, FL must navigate the balance between preserving previous global knowledge and acquiring new local knowledge during local learning. Inspired by CL, we aim to resolve the conflict between two objectives through \textit{Orthogonal Learning}\,\citep{ogd, gem}. In CL, an effective strategy for preserving previous knowledge is to minimizing the new task's loss using gradients that are orthogonal to the loss space of the old tasks\,\citep{gpm, agem, cl_low_rank, cl_rethinking_gradient_projection}. These orthogonal gradients enable the model to reduce the loss on the new task without negatively affecting performance on previous tasks. Our primary motivation is to adapt this strategy to FL by identifying gradients that are orthogonal to directions affecting the global knowledge while still enabling effective training on local datasets.

However, applying this strategy to FL presents unique challenges. First, implementing orthogonal gradients in CL often requires retaining past data or gradients for reference\,\citep{gpm, agem}. This practice becomes problematic in FL as it can compromise data privacy and introduce additional communication overhead. Second, unlike in CL where task distributions are often disjoint\,\citep{owm, fs_dgpm}, FL clients may have overlapping data distributions, including instances from the same class. As a result, the global distribution, formed by combining local distributions, also presents overlap with individual local distributions. Such an overlap not only complicates the identification of an orthogonal update direction but also makes the process computationally demanding. Moreover, finding an orthogonal gradient that accommodates multiple local distributions is challenging and the obtained gradient may significantly undermine the effectiveness of learning on local datasets. 

To address these problems, we initially focus on \textit{proximal restrictions} in FL\,\citep{scaffold, moon, fedprox}. Many previous studies have integrated these restrictions into local objectives to tackle the data heterogeneity issue. By constraining the deviation of local learning from the global objective, the proximal restrictions maintain performance on the global distribution outside of local distributions. This leads us to interpret proximal objectives as losses that preserve global knowledge in FL, notably without requiring direct access to other clients' data or the need to communicate their information. Unfortunately, as the proximal objective is closely tied to the local objective, we find that directly negating the projected proximal gradient component in the local gradient substantially degrades the performance.

Instead, we hypothesize that identifying the parameter region where local gradients naturally align as orthogonal to the proximal gradients can effectively reduce interference between the two objectives. To this end, we propose a novel algorithm, Federated Stabilized Orthogonal Learning\,(FedSOL). At each local update, FedSOL adversarially perturbs weight parameters using the gradient of proximal objectives, and then captures the local gradient at these perturbed weights. FedSOL aims to find a parameter region where the local gradient is stable against proximal perturbation, ensuring that the resulting local gradient remains orthogonal to the proximal gradient.

\vspace{2pt}
\noindent To summarize, our main contributions are as follows:

\begin{itemize}
    \item We suggest that orthogonal learning in CL could be an effective strategy in FL, by resolving conflicts between local and proximal objectives. \textbf{(\Autoref{sec:motiv})}
    
    \item We propose a novel FL method, FedSOL, which targets a parameter region where the local gradient is orthogonal to the proximal gradient. \textbf{(\Autoref{sec:method})}
    
    \item We validate the efficacy of FedSOL in various setups, demonstrating consistent state-of-the-art performance. We also highlight its robustness when integrated with different types of proximal objectives. \textbf{(\Autoref{sec:experiment})}
    
    \item We provide a comprehensive analysis of the benefits that FedSOL offers to FL, effectively preserving global knowledge during local learning and enhancing the smoothness of the global model. \textbf{(\Autoref{sec:analysis})}
\end{itemize}

%% file: sec/02-motiv.tex
\section{Proximal Restriction in Local Learning}
\label{sec:motiv}
\vspace{-2pt}
In this section, we first introduce the problem setup and the concept of proximal restriction in FL. We then discuss the trade-off between local and proximal objectives. We suggest that while orthogonal learning could be an effective solution, simple gradient projection cannot achieve it in FL.

\subsection{Proximal Restriction in FL}
\vspace{-1.5pt}
Consider an FL system that consists of $K$ clients and a central server. Each client $k$ has a local dataset $\mathcal{D}^k$, where the entire dataset is a union of the local datasets as $\mathcal{D} = {\bigcup}_{k\in[K]}\mathcal{D}^k$. FL aims to train a global server model with weights $\bw$ that minimize the loss across all clients:
\begin{equation}
    \vspace{-1.5pt}
    \mathcal{L}_{\mathrm{global}}(\bw) = \underset{{k\in[K]}}{{\sum}}\, \frac{|\mathcal{D}_k|}{|\mathcal{D}|}\,\mathcal{L}_{\mathrm{local}}^k(\bw)\,,
\end{equation}

\noindent
where $|\mathcal{D}^k|$ and $|\mathcal{D}|$ are the number of instances in each dataset. When using a proximal restriction objective, the loss function for each client $k$ is a linear combination of its original local loss, $\mathcal{L}_{\mathrm{local}}^k(\bwk)$, and a proximal loss, $\mathcal{L}_p^k(\bwk;\bwg)$, controlled by a hyperparameter $\beta$:
\begin{equation}
    \mathcal{L}^k(\bwk) = \mathcal{L}_{\mathrm{local}}^k(\bwk) + \beta\cdot\mathcal{L}_p^k(\bwk;\bwg).
    \label{eq_proximal_restriction}
\end{equation}
Here, $\mathcal{L}_{\mathrm{local}}^k(\bwk)$ is the loss on the client's local distribution\,(e.g., cross-entropy loss), and $\mathcal{L}^k_p(\bwk;\bwg)$ quantifies the discrepancy between the global model $\bwg$ and the local model $\bwk$. This discrepancy can be measured in various ways, such as the Euclidean distance between the parameters\,\citep{fedprox, fednova} or the KL-divergence between probability vectors computed using the client's data\,\citep{feddistill, fedntd}.

\subsection{Forgetting in Local Learning}
\vspace{-1.5pt}
Recent studies suggest that data heterogeneity in FL leads to \textit{Catastrophic Forgetting} during local learning\,\citep{fedcurv, fedntd}. When the model is trained on skewed local datasets, local learning deviates from the global objectives—commonly referred to as client drift\,\citep{scaffold}. This drift causes trained local models to forget knowledge from the previous round of the global model, which the local datasets cannot fully represent. The FL performance is strongly tied to how well local learning preserves this global knowledge\,\citep{fedreg, forgetting_survey}. Introducing proximal restrictions within the local objective effectively constrains such deviation, alleviating forgetting\,\citep{fedntd}.

However, these proximal restrictions present a trade-off. While they serve to preserve global knowledge, they also inherently limit the model's ability to learn from local data\,\citep{fedsam, fedalign}. Striking the right balance between these two conflicting objectives during local learning is crucial for the success of FL. As the local model $\bwk$ begins with the same parameters as the distributed global model $\bwg$, it initially has a minimal proximal loss. Thereby, we posit that the main challenge is guiding the local learning to reduce the local loss $\mathcal{L}^k_{\mathrm{local}}$ without inducing an increase in the proximal loss $\mathcal{L}^k_{p}$. Inspired by CL\,\citep{agem, gpm, ogd}, we consider updating the local model using gradients that are orthogonal to the proximal gradient as an effective solution to this problem.

\subsection{Proximal Gradient Projection}
\vspace{-1.5pt}
A straightforward approach to obtaining the update gradient, which is orthogonal to the proximal loss $\mathcal{L}_p^k$, is conducting a direct projection\,\citep{agem, gpm} of the proximal gradient:
\begin{equation}
    \vspace{-1pt}
    \boldsymbol{g}_u^{\text{Proj}} = \boldsymbol{g}_l - \frac{{\boldsymbol{g}_l}^{T}\;\boldsymbol{g}_p}{\boldsymbol{g}_{p}^{T}\;\boldsymbol{g}_{p}}\boldsymbol{g}_p\qquad \text{if}\quad \boldsymbol{g}_l^{T} \; \boldsymbol{g}_p < 0.
    \label{eq_proj}
\end{equation}

\noindent Here, we denote the local gradient as $\boldsymbol{g}_l=\nabla_{\bwk}\mathcal{L}^k_{\mathrm{local}}(\bwk)$ and the proximal gradient as $\boldsymbol{g}_p= \nabla_{\bwk} \mathcal{L}^k_{p}(\bwk;{\bwg})$. We omit the weights $\boldsymbol{w}$ for simplicity unless clarification is needed. By negating the conflicting component from the local gradient $\boldsymbol{g}_l$, the update gradient $\boldsymbol{g}_u^{\text{Proj}}$ becomes orthogonal to the proximal gradient $\boldsymbol{g}_p$. Note that we only project when the two gradients are in conflict (i.e., $\boldsymbol{g}_l^{T}\;\boldsymbol{g}_p < 0$), otherwise we use the original local gradient $\boldsymbol{g}_l$.

However, we find that this direct projection approach rather degrades the performance. In \autoref{tab:grad_proj}, we compare different usages of the proximal objective: as an auxiliary loss alongside the local objective, as in \autoref{eq_proximal_restriction} (Base), and as proximal gradient projection, as in \autoref{eq_proj} (Projection). Note that the absence of a proximal loss (None) is equivalent to FedAvg\,\citep{fedavg}. The results show that performance significantly declines when the proximal gradient component is directly negated through gradient projection, even underperforming compared to FedAvg. This suggests that the two objectives are closely interconnected, and directly negating the proximal gradient might actually undermine the local learning. Therefore, we consider an approach that implicitly promotes the orthogonality of the update.
\vspace{-4pt}
\begingroup
\setlength{\tabcolsep}{10pt} 
\renewcommand{\arraystretch}{1.08}
\begin{table}[ht!]
\small
\caption{Results of gradient projection on CIFAR-10 ($\alpha$ = 0.1).}
\label{tab:grad_proj}
\vspace{-4pt}
\centering
\begin{tabular}{lcc} 
\toprule
\multicolumn{1}{c}{\multirow{2}{*}{\textbf{Proximal Loss }}} & \multicolumn{2}{c}{\textbf{Usage}}  \\
\multicolumn{1}{c}{}                                         & Base  & Projection                   \\ 
\hline\hline
None (FedAvg)                                                         &\multicolumn{2}{c}{56.13$_{\pm 0.78}$}  \\                  
\hline
L2 Distance                                                           & \textbf{59.80}$_{\pm 1.12}$ & 56.35$_{\pm 2.85}$ \textcolor{red!90!black}{(- 3.45)}                        \\
KL-Divergence                                                       & \textbf{60.31}$_{\pm 2.07}$& 50.88$_{\pm 3.55}$ \textcolor{red!90!black}{(- 9.43)}\\
\bottomrule
\end{tabular}
\end{table}
\endgroup

%% file: sec/03-method.tex
\section{Proposed Method: FedSOL}
\label{sec:method}
\vspace{-1pt}

In this section, we introduce Federated Stabilized Orthogonal Learning\,(FedSOL). Our primary motivation is to obtain a gradient for updating the local model that is orthogonal to the proximal gradient, yet still effectively reduces the local loss. The detailed algorithm is outlined in \Autoref{algo:fedsol}.

\subsection{Preliminary: Overview of SAM} We borrow the idea of recently proposed Sharpness-Aware Minimization\,(SAM)\,\citep{sam}, which uses weight perturbations to achieve flatter minima. For the given loss $\mathcal{L}$, SAM optimizer solves the following min-max problem:
\begin{equation}
    \underset{\bw}{\text{min}} \; \underset{{\| \boldsymbol{\epsilon} \|}_{2} < \rho }{\text{max}} \mathcal{L}(\bw+\boldsymbol{\epsilon}). 
    \label{eq_sam_minmax}
\end{equation}
In the above equation, the inner maximization identifies a parameter perturbation $\epsilon$ that maximizes loss change within the $\rho$-ball neighborhood. This is practically approximated by a single re-scaled gradient step ${\boldsymbol{\epsilon}}^{*}$$=$$\rho {\nabla}_{\bw}\mathcal{L}(\bw)/\|{\nabla}_{\bw}\mathcal{L}(\bw)\|_2$. The outer minimization is then conducted by a base optimizer, such as SGD\,\citep{sgd}, taking the gradient ${\nabla}_{\bw}\mathcal{L}(\bw + {\boldsymbol{\epsilon}}^{*})$ at the perturbed weights. SAM demonstrates an exceptional ability to perform well across different model structures\,\citep{sam_on_vit, gsam} and tasks\,\citep{sam_domain_generalization, sam_maml} with high generalization performance. In FL, applying SAM improves the generalization of each client's local model\,\citep{fedsam, fedasam}. However, since this approach only addresses local objectives, its effectiveness in generalization is mostly confined to local data distributions\,\citep{fedsmoo, fedspeed} and still encounter conflicts with the proximal objective.

\subsection{Adversarial Proximal Perturbation}

\begin{figure}[ht!]
    \centering
    \includegraphics[width=0.478\textwidth]{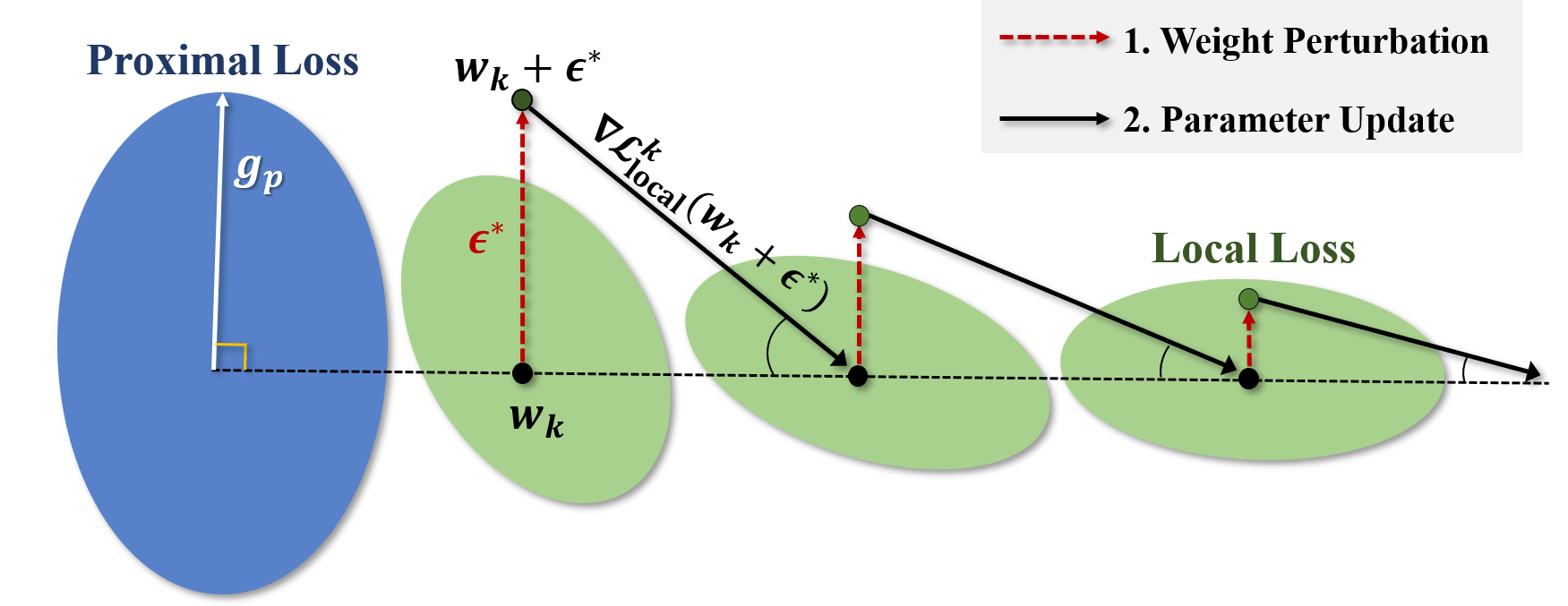}
    \caption{An overview of the FedSOL update. At each update, FedSOL computes its update gradient at the proximally perturbed weights. By withstanding the proximal perturbation, FedSOL obtains a local gradient that is orthogonal to the proximal gradient.}
    \label{fig:fedsol_update}
    \vspace{-10pt}
\end{figure}

\input{materials/algorithm}
\noindent The core idea of FedSOL is to identify the parameter region where the local gradient is minimally affected by adversarial weight perturbation using the proximal gradient. At each local update, FedSOL adversarially perturbs the weight parameters and obtains the local gradient as follows:

\paragraph{Step1: Weight Perturbation} In FedSOL, we first identify a weight perturbation $\boldsymbol{\epsilon}^{*}_p$ that brings about the most significant change in the given proximal loss $\mathcal{L}^k_p$. Practically, this is conducted by using a single re-scaled step based on the proximal gradient $\boldsymbol{g}_p= \nabla_{\bwk} \mathcal{L}^k_{p}(\bwk;{\bwg})$, controlled by the hyperparameter $\rho$ for perturbation strength:
\begin{equation}
\boldsymbol{\epsilon}^{*}_p =\; \underset{\|\boldsymbol{\epsilon}\|_2 \leq \rho}{\text{argmax}}\; \mathcal{L}^k_{p}(\bwk \!+\! \boldsymbol{\epsilon};\bwg) \;\approx\; \rho\,\frac{\boldsymbol{g}_p}{\|\boldsymbol{g}_p\|_2}  \,.
\label{eq_fedsol_epsilon}
\end{equation}

\paragraph{Step2: Parameter Update} After the perturbation, update the parameters with the local gradient computed at the perturbed weights and a learning rate $\gamma$:
\begin{equation}
    \bwk \leftarrow \bwk - \gamma \cdot {\nabla}_{\bwk}\mathcal{L}^k_{\mathrm{local}}(\bwk + \boldsymbol{\epsilon}^{*}_p)\,,
    \label{eq_fedsol_update}
\end{equation}

In the above procedures, the model is updated using the gradient of the local loss $\mathcal{L}^k_{\mathrm{local}}$, while the proximal loss $\mathcal{L}_{{p}}^k$ plays only an implicit role in perturbing weights. 
We emphasize that unlike SAM, which employs the same loss for both weight perturbations and parameter updates, FedSOL distinguishes between the roles of these two types of losses to address the knowledge trade-off in FL. A more detailed comparison of these two distinct approaches in the FL context is discussed in \autoref{appendix_discussion_on_sam}.

\subsection{Adaptive Perturbation Strength}
\label{subsec:adaptive_perturbation}
In FedSOL, we introduce an adaptive perturbation strength reflecting the global and local parameter discrepancies. For each layer $m$, we construct a scaling vector $\boldsymbol{\lambda}^{(m)}$, where the $i$-th entry corresponds to each parameter in that layer:
\begin{equation}
\label{eq:adaptive_perturbation}
\boldsymbol{\lambda}^{(m)}[i] = \frac{\big|{\bw}^{(m)}_{k}[i] - {\bw}^{(m)}_{g}[i]\big|}{\big\| {\bw}^{(m)}_{k} - {\bw}^{(m)}_{g} \big\|_2}.
\end{equation}

Here, $\bwg^{(m)}$ and $\bwk^{(m)}$ represent the weights of the $m$-th layer in the global and local models, respectively. The denominator represents the normalization of the discrepancy within the layer, accounting for the layer-wise scale variance. This adaptive perturbation allows more perturbation for the parameter with large difference, and vice versa. It fits with the typical behavior of proximal loss, which increases as $\|\bwk - \bwg\|_2$ grows. By concatenating these layer-specific vectors, $\boldsymbol{\Lambda} = (\boldsymbol{\lambda}^{(1)},\ldots, \boldsymbol{\lambda}^{(m)}, \ldots, \boldsymbol{\lambda}^{(\mathrm{last})})$, and incorporating it into the \Autoref{eq_fedsol_epsilon}, the proximal perturbation $\boldsymbol{\epsilon}^{*}_p$ becomes:
\begin{equation}
{\boldsymbol{\epsilon}}^{*}_p 
= \:\: \rho \cdot \boldsymbol{\Lambda} \odot \frac{\boldsymbol{g}_p}{\|\boldsymbol{g}_p\|_2} \approx \underset{\left\lVert \boldsymbol{\Lambda} ^{-1} \!\odot\,\boldsymbol{\epsilon}\right\rVert_2 \leq \rho}{\text{argmax}}\; \mathcal{L}^k_{p}(\bwk + \boldsymbol{\epsilon};\bwg)\,,
\label{eq_adaptive_sol}
\end{equation}

\noindent where $\odot$ denotes the element-wise product. Intuitively, the adaptive perturbation allows local learning to deviate certain parameters from the global model when they are significantly influential to justify larger weight perturbations. Note that using a fixed value for $\rho$ corresponds to setting $\boldsymbol{\Lambda}$ in \Autoref{eq_adaptive_sol} as a vector with all entries to one.

\subsection{Partial Perturbation}

As the data heterogeneity does not affect all layers equally\,\citep{ccvr, etf}, we explore the use of \textit{partial} perturbation in FedSOL by selectively perturbing specific layers instead of the entire model. We propose that perturbing only the last classifier layer is empirically sufficient for FedSOL. We provide a detailed discussion in \Autoref{sec4:ablation_study}. This approach yields performance nearly as high as full-model perturbation while significantly reducing computational requirements by avoiding multiple forward and backward computations across all layers. For optimal efficiency and performance, we apply FedSOL's weight perturbation exclusively to the last classifier head layer in the experiments unless otherwise specified.

\subsection{Theoretical Analysis}
To assess FedSOL's effect on local learning, we examine the changes in both $\mathcal{L}^k_{\mathrm{p}}$ and the local loss $\mathcal{L}^k_{\mathrm{local}}$ after a single FedSOL update. Note that the proximal loss $\mathcal{L}^k_{\mathrm{p}}$ and local loss $\mathcal{L}^k_{\mathrm{local}}$ each corresponds to the global knowledge and local knowledge. Specifically, we examine the impact of the FedSOL update gradient $\boldsymbol{g}_u^{\text{FedSOL}}$ on loss $\mathcal{L}^k$ at $\bwk$ with a learning rate $\gamma$, using a first-order Taylor approximation:

\begin{flalign}
\Delta^{\mathrm{FedSOL}}\mathcal{L}^k(\bwk) 
&= \mathcal{L}^k\big(\bwk - \gamma\,\boldsymbol{g}_u^{\text{FedSOL}}(\bwk)\big) - \mathcal{L}^k(\bwk)\notag \\
&\approx -\gamma \langle \nabla_{\bwk}  \mathcal{L}^k(\bwk),\; \boldsymbol{g}_u^{\text{FedSOL}}(\bwk)\rangle
\,.
\label{eq_loss_difference_definition}
\vspace{2pt}
\end{flalign}
In the above equation, $\mathcal{L}^k$ can be either local loss or proximal loss. We further scrutinize $\boldsymbol{g}_u^{\text{FedSOL}}$ itself as a first-order Taylor expansion of the local loss $\mathcal{L}^k_{\mathrm{local}}$ at the perturbed weights $\bwk + \boldsymbol{\epsilon}^{*}_p$ as follows:
\begin{flalign}
\boldsymbol{g}_u^{\mathrm{FedSOL}}(\bwk)&={\nabla}_{\bwk} \mathcal{L}^k_{\mathrm{local}}(\bwk+\boldsymbol{\epsilon}^{*}_p) \notag \\
& \approx \boldsymbol{g}_l(\bwk) + \rho\, \nabla^2_{\bwk} \mathcal{L}^k_{\mathrm{local}}(\bwk)\hat{\boldsymbol{g}}_p(\bwk)\,.
\label{eq_sol_taylor}
\vspace{2pt}
\end{flalign}
where $\hat{\boldsymbol{g}}_p$$=$$\boldsymbol{g}_p/\|\boldsymbol{g}_p\|_2$ represents the normalized proximal gradient. Note that $\boldsymbol{\epsilon}^{*}_p$ is solely used for weight perturbation, and hence its gradient is not computed. By integrating the approximation in \Autoref{eq_sol_taylor} into the loss difference defined in \Autoref{eq_loss_difference_definition}, we derive the subsequent two key propositions. These propositions explain how FedSOL guides local learning to minimize the local loss, $\mathcal{L}^k_{\mathrm{local}}$, without causing an increase in the proximal loss, $\mathcal{L}^k_{\mathrm{p}}$.

\begin{proposition}
\textup{(Proximal Objective Orthogonality).}\; Given a local loss $\mathcal{L}^k_{\mathrm{local}}$ and its Hessian matrix $\nabla^2\mathcal{L}^k_{\mathrm{local}} \succcurlyeq 0$ evaluated at $\bwk$, the change of proximal loss by FedSOL update reduces the conflicts $\langle\boldsymbol{g}_l$\,,$\boldsymbol{g}_p\rangle \leq 0$ in FedAvg update $\Delta^{\mathrm{FedAvg}}\mathcal{L}^k_{p}=-\gamma\,\langle \boldsymbol{g}_l\,,\boldsymbol{g}_p\rangle$ as $\rho$ increases:
\begin{equation}
\Delta^{\mathrm{FedSOL}} \mathcal{L}^k_{p} \approx -\gamma\Big(\langle \boldsymbol{g}_l\,,\boldsymbol{g}_p \rangle + \rho \cdot \zzunderbrace{\hat{\boldsymbol{g}}_p{\!\!}^\top \nabla^2\mathcal{L}^k_{\mathrm{local}}\,{\boldsymbol{g}_p}}{$\geq 0$}\Big)\,.
\label{eq_loss_difference}
\end{equation}
\label{prop1:proximal_change}
\vspace{-3pt}

\end{proposition}
\begin{proposition}
\textup{(Local Objective Equivalence).}\; The change of local loss $\mathcal{L}^k_{\mathrm{local}}$ by FedSOL update is equivalent to the FedAvg update conducted at $\nabla \mathcal{L}^k_{\mathrm{local}}(\bwk + \frac{\rho}{2}\boldsymbol{\epsilon}^{*}_p)$ as:
\begin{equation}
\Delta^{\mathrm{FedSOL}} \mathcal{L}^k_{\mathrm{local}}(\bwk) \approx \Delta^{\mathrm{FedAvg}} \mathcal{L}^k_{\mathrm{local}}\Big(\bwk + \frac{\rho}{2}\boldsymbol{\epsilon}^{*}_p \Big)\,.
\end{equation}
\label{prop2:local_change}
\vspace{-6pt}
\end{proposition}

Firstly, \textbf{Proposition 1} examines FedSOL's impact on the proximal loss. It suggests that FedSOL's update gradient, $\boldsymbol{g}_u^{\text{FedSOL}}$, directs the local updates to be orthogonal to the proximal gradient. This helps maintain a low proximal loss, $\mathcal{L}^k_{\mathrm{p}}$, during local learning, which initially has a very low value as the learning starts from the distributed global model. This indicates that FedSOL implicitly regularizes the negative impact of the local gradient on proximal loss. This regularization effect grows as the curvature of local loss $\nabla^2\mathcal{L}^k_{\mathrm{local}}$ local becomes steeper.

Meanwhile, \textbf{Proposition 2} compares the change of local loss $\mathcal{L}^k_{\mathrm{local}}$ under FedSOL to its counterpart in FedAvg. This proposition suggests that, although FedSOL calculates the local gradient at perturbed weights, its impact on the local loss is almost identical to that of FedAvg. This implies that FedSOL effectively reduces local loss without significantly slowing down the learning process. As a result, FedSOL successfully mitigates the conflict between the proximal objective and the local objective. The detailed proofs of the propositions are provided in \Autoref{appendix_proof}.

%% file: materials/algorithm.tex
\newcommand*{\tikzmk}[1]{\tikz[remember picture,overlay,] \node (#1) {};\ignorespaces}
%
\newcommand{\boxit}[1]{\tikz[remember picture,overlay]{\node[yshift=3pt,fill=#1,opacity=.20,fit={(A)($(B)+(.95\linewidth,.8\baselineskip)$)}] {};}\ignorespaces}
\begin{algorithm}[t!]
    \caption{\textbf{Fed}erated \textbf{S}tabilized \textbf{O}rthogonal \textbf{L}earning}
    \textbf{Input:} local loss $\mathcal{L}^k_{\mathrm{local}}$ and proximal loss $\mathcal{L}^k_{p}$ for each client $k\in[K]$, learning rate $\gamma$, and base perturbation strength $\rho$
    \par
    \textbf{Initialize} global server weight $\boldsymbol{w}_g$\par
    \textbf{for} each communication round $t$ \textbf{do} \par
    \quad Server samples clients $K^{(t)} \subset [K]$ \par
    \quad Server broadcasts ${\boldsymbol{w}_g}$ for all $k\in K^{(t)}$\par
    \quad Client replaces ${\boldsymbol{w}_k}\leftarrow{\boldsymbol{w}_g}$ \par
    \quad \textbf{for} each client $k \in K^{(t)}$ \textbf{in parallel do} \par
    \quad \quad \textbf{for} each local step \textbf{do} \par
    \vspace{1pt}
    \tikzmk{A}
    \vspace{2pt}
    \!\quad \quad \quad \textcolor{red!90!black}{\# Set Adaptive Perturbation Radius (Sec 3.3)}
    \par
    \quad \quad \quad $\boldsymbol{\rho}_{\mathrm{adaptive}} = \rho \cdot \boldsymbol{\Lambda}$ \quad (\textit{element-wise rescale})
    \par
    \vspace{7pt}
    \quad \quad \quad \textcolor{red!90!black}{\# Perturb using Proximal Gradient (Sec 3.2)}
    \par
    \quad \quad \quad \textcolor{red!90!black}{\# (Optional) Use Partial Perturbation (Sec 3.3)}
    \par
    \quad \quad \quad $ {\boldsymbol{\epsilon}}^{*}_p = \boldsymbol{\rho}_{\mathrm{adaptive}} \odot 
    \frac{\nabla_{\boldsymbol{w}_k}{\mathcal{L}^k_{p}(\boldsymbol{w}_k;\boldsymbol{w}_g)}} {\| \nabla_{\boldsymbol{w}_k}{\mathcal{L}^k_{p}(\boldsymbol{w}_k;\boldsymbol{w}_g)} \|} $
    \par
    \vspace{7pt}
    \quad \quad \quad \textcolor{red!90!black}{\# Update Local Model Parameters (Sec 3.2)}
    \par
    \quad \quad \quad ${\boldsymbol{w}_k} \leftarrow {\boldsymbol{w}_k} - \gamma \cdot \nabla_{\boldsymbol{w}_k} \mathcal{L}^k_{\mathrm{local}}\big(\boldsymbol{w}_k + {\boldsymbol{\epsilon}}^{*}_p\big)$
    \vspace{4pt}
    \par
    \tikzmk{B}
    \boxit{blue!60!gray}
    \quad \quad \textbf{end for} \par
    \quad \textbf{end for} \par
    \quad Upload ${\boldsymbol{w}_k}$ to server \par
    \quad \textbf{Server Aggregation :}$ {\boldsymbol{w}_g} \leftarrow \frac{1}{|K^{(t)}|}\sum_{k \in K^{(t)}} {\boldsymbol{w}_k}$ \par
    \textbf{end for} \par 
    \textbf{Server output :} ${\boldsymbol{w}_g}$
    \label{algo:fedsol}
\end{algorithm}

%% file: sec/04-experiment.tex
\section{Experiment}
\label{sec:experiment}
\vspace{-1pt}

\subsection{Experimental Setups}

\begin{figure}[ht!]
    \centering
\includegraphics[width=0.475\textwidth]{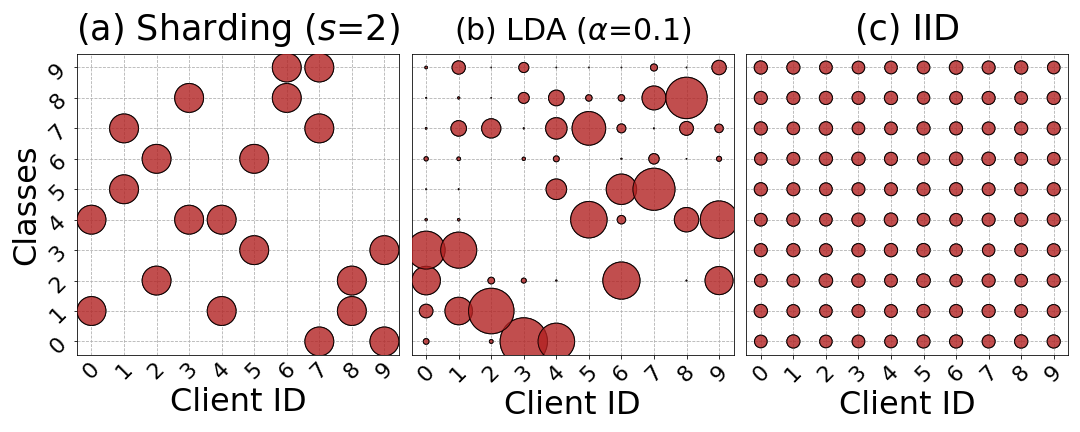}
    \vspace{-18pt}
    \caption{CIFAR-10 partition examples across 10 clients.}
    \label{fig:partition}
    \vspace{-4pt}
\end{figure}

\paragraph{Data Setups} We use 6 datasets: MNIST\,\citep{mnist}, CIFAR-10\,\citep{cifar}, SVHN\,\citep{svhn}, CINIC-10\,\citep{cinic10}, PathMNIST\,\cite{medmnist}, and TissueMNIST\,\citep{medmnist}. We distribute data to clients using two strategies: Sharding\,\citep{fedavg, fedbabu} and Latent Dirichlet Allocation\,(LDA)\,\citep{fedma, fedntd}. Sharding sorts data by label and assigns equal-size shards to clients. The heterogeneity level increases as the shard per user, $s$, becomes smaller. On the other hand, LDA assigns class $c$ data samples to each client $k$ with probability $p_c (\approx \text{Dir}(\alpha))$, where the heterogeneity increases as $\alpha$ becomes smaller. \autoref{fig:partition} illustrates the differences between these two partition strategies. Note that although only the statistical distributions varies across the clients in Sharding strategy, both the distribution and dataset size differ in LDA strategy.

\paragraph{Learning Setups}  We distribute MNIST, CIFAR-10, and SVHN datasets across 100 clients with a sampling ratio of 0.1, while CINIC-10, PathMNIST, and TissueMNIST across 200 clients with a ratio of 0.05. We use a model architecture as described in\,\citep{fedavg}, which consists of two convolutional layers, max-pooling layers, and two fully connected layers. Each client optimizes its local datasets for 5 local epochs using momentum SGD with a learning rate of 0.01, momentum 0.9, and weight decay 1e-5. The learning rate is multiplied by a factor of 0.99 after each communication round. We conducted a total of 300 communication rounds, except for MNIST, PathMNIST, and TissueMNIST, for which we conducted 200 rounds, sufficient for the server model to reach performance saturation. For the proximal loss, we employed the KL-divergence loss function. We provide more detailed experimental setups in \autoref{appendix_exeprimental_setups}.

\vspace{2pt}
\subsection{Proximal Orthogonality of FedSOL}
\begin{figure}[ht!]
    \centering
    \includegraphics[width=0.475\textwidth]{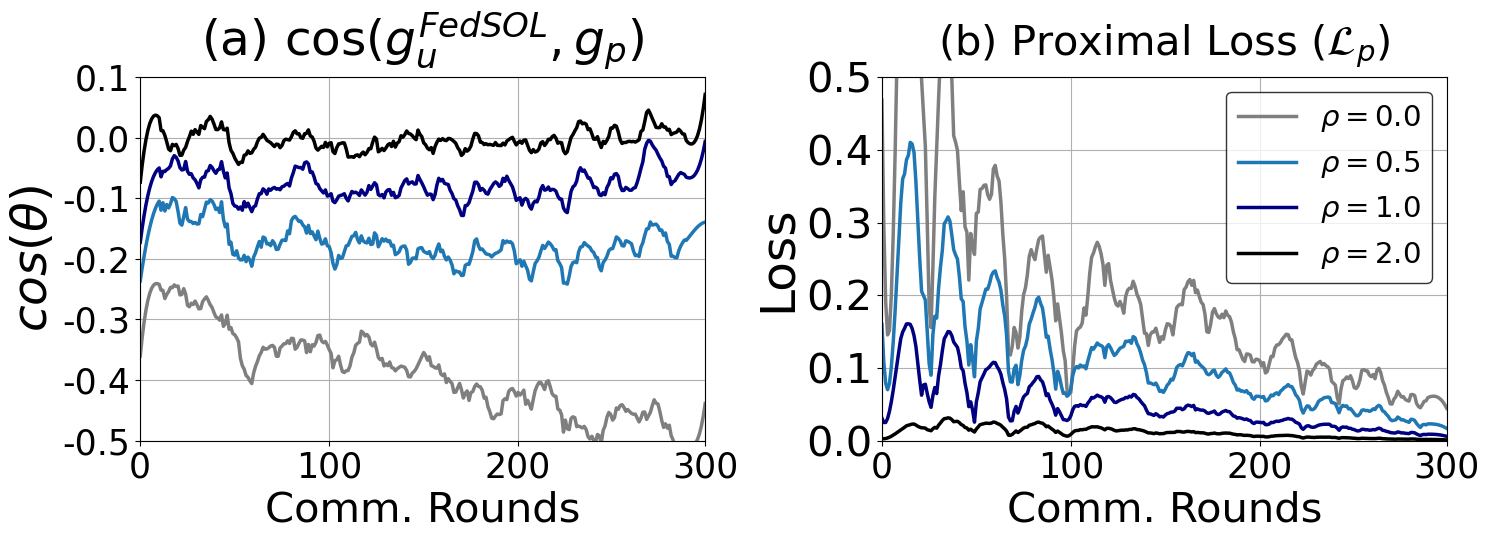}
    \vspace{-12pt}
    \caption{Effect of FedSOL on local learning in CIFAR-10 ($\alpha$=0.1) by varying $\rho$ values. (a) Average proximal loss across local models. (b) Cosine similarity between FedSOL gradient ($g_u^{\text{FedSOL}}$) and proximal gradient ($g_p$) during local learning.}
    \label{fig:fedsol_analysis}
    \vspace{-6pt}
\end{figure}

\vspace{-1pt}
In \autoref{fig:fedsol_analysis}, we examine  the interaction of FedSOL's update gradient, $g_u^{\text{FedSOL}}$, with the proximal loss $\mathcal{L}_p$. As the perturbation strength $\rho$ increases, the direction of $g_u^{\text{FedSOL}}$ becomes increasingly orthogonal to the proximal gradient $g_p$ (\autoref{fig:fedsol_analysis}\textcolor{red}{(a)}). This enhanced orthogonality helps maintain a low proximal loss during local learning (\autoref{fig:fedsol_analysis}\textcolor{red}{(b)}).

\input{materials/tables/main_exp}

\subsection{Performance on Data Heterogeneity}
\vspace{-1pt}
\paragraph{Heterogeneity Level} \Autoref{tab:main_std} presents a comparison between our approach, FedSOL, and other baselines. The results show that many recently proposed FL methods tend to underperform even when compared to the standard FedAvg baseline. A similar observation is also reported in\,\citep{fl_with_noniid, fedntd}, which points out that many FL methods are sensitive to the learning scenarios. In contrast, FedSOL achieves state-of-the-art results in most cases, consistently outperforming FedAvg across all evaluated scenarios. Particularly, FedSOL shows remarkable improvement at high heterogeneity levels, such as in Sharding\,($s$=2) and LDA\,($\alpha$=0.05) scenarios. We further provide the learning curves in \autoref{appendix_learning_curves}, comparison to other weights perturbation strategies in \autoref{appendix_perturb_strategy}, personalized FL performance in \autoref{appendix_pfl_performance}, and performance on larger datasets in \autoref{appendix_cifar100_imagenet100}.
\vspace{-10pt}
\begin{figure*}[t!]
    \vspace{-1pt}
    \centering
    \includegraphics[width=0.94\textwidth]{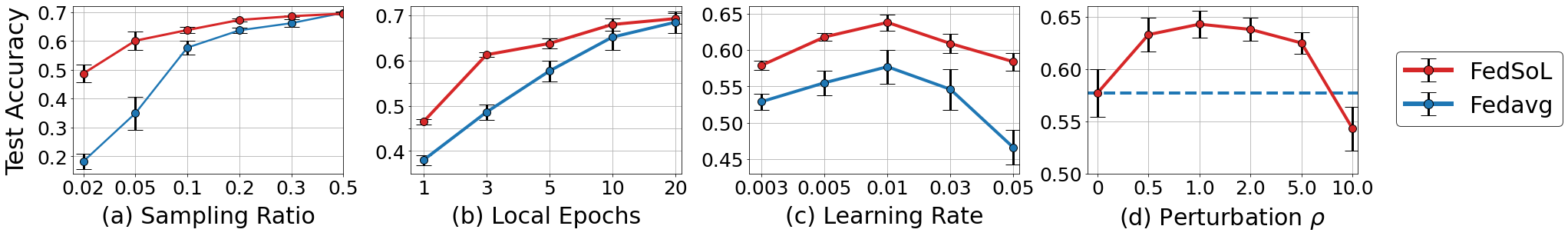}
    \vspace{-4pt}
\caption{{Performance of \textit{FedAvg} and \textit{FedSOL} on CIFAR-10 ($\alpha$=0.1) with various setups: (a) sampling ratio, (b) the number of local epochs, (c) initial learning rate, and (d) perturbation strength. The error bars stand for the standard deviations.}}
\vspace{-8pt}
\label{fig:ablation}
\end{figure*}

\paragraph{Learning Factors} We examine the learning factors that influence FedSOL's performance in \autoref{fig:ablation}. Note that the sampling of clients simulates communication failures with inactive devices. In our experiments, FedSOL consistently surpasses FedAvg across various factors, achieving its best performance within the $\rho$ range between 0.5 and 2.0. Most notably, FedSOL's gains increase as a smaller portion of clients participate in each round. For example, FedAvg's performance significantly declines at a sampling ratio of 0.02, falling to near-random accuracy. However, FedSOL remains robust under such conditions. Further comparisons are provided in \autoref{appendix_learning_factors_fedsam_fedasam}.
\vspace{-5pt}

\paragraph{Model Architecture} We conduct further experiments using different model architectures: VggNet-11\,\citep{vggnet}, ResNet-18\,\citep{resnet}, and SL-ViT\,\citep{vit_small_datasets}, which is a specialized structure of ViT\,\citep{vit} designed for small-sized datasets. The results provided in \autoref{tab:ablation_architecture} validate the robust efficacy of FedSOL across various model architectures.

\begingroup
\setlength{\tabcolsep}{4.5pt} 
\renewcommand{\arraystretch}{1.165}
\begin{table}[ht!]
\vspace{-1pt}
\caption{Comparison of methods on different model architectures. The heterogeneity is set as LDA ($\alpha$ = 0.1).}
\vspace{-2pt}
\centering
\small
\begin{tabular}{clccc} 
\toprule
\textbf{Model}          & \multicolumn{1}{c}{\textbf{Method}} & \textbf{CIFAR-10} & \textbf{SVNH} & \textbf{PathMNIST}  \\ 
\hline\hline
\multirow{4}{*}{Vgg11}  
& FedAvg                              
& 41.30${}_{\pm 1.07}$             
& 50.02${}_{\pm 4.25}$         
& 61.79${}_{\pm 9.88}$              \\
& FedProx                            
& 40.45${}_{\pm 1.41}$             
& 31.07${}_{\pm 6.72}$         
& 63.47${}_{\pm 2.68}$              \\
& FedNTD                              
& \underline{\textbf{60.55}}${}_{\pm 2.14}$             
& 56.62${}_{\pm 2.64}$         
& 69.82${}_{\pm 2.27}$              \\
& \textbf{FedSOL}                           
& 56.39${}_{\pm 1.40}$             
& \underline{\textbf{74.74}}${}_{\pm 0.04}$             
& \underline{\textbf{78.38}}${}_{\pm 1.12}$             
\\ 
\hline
\multirow{4}{*}{Res18}  
& FedAvg                              
& 49.92${}_{\pm 0.62}$             
& 76.98${}_{\pm 2.90}$         
& 57.91${}_{\pm 1.27}$              \\
& FedProx                            
& 59.00${}_{\pm 2.58}$             
& 82.09${}_{\pm 2.35}$         
& 75.84${}_{\pm 1.58}$              \\
& FedNTD                              
& 57.79${}_{\pm 3.42}$             
& 78.50${}_{\pm 0.18}$         
& 76.87${}_{\pm 0.57}$              \\
& \textbf{FedSOL}                           
& \underline{\textbf{66.32}}${}_{\pm 0.48}$             
& \underline{\textbf{85.97}}${}_{\pm 0.04}$             
& \underline{\textbf{80.59}}${}_{\pm 0.11}$             
\\ 
\hline
\multirow{4}{*}{SL-ViT} 
& FedAvg                              
& 35.48${}_{\pm 2.09}$             
& 53.94${}_{\pm 5.17}$         
& 72.44${}_{\pm 1.91}$              \\
& FedProx                            
& 38.73${}_{\pm 1.23}$             
& 58.25${}_{\pm 4.23}$         
& 74.10${}_{\pm 1.23}$              \\
& FedNTD                              
& 47.59${}_{\pm 2.84}$             
& 61.46${}_{\pm 1.76}$         
& 71.65${}_{\pm 1.71}$              \\
& \textbf{FedSOL}                           
& \underline{\textbf{47.95}}${}_{\pm 1.51}$             
& \underline{\textbf{67.19}}${}_{\pm 0.33}$             
& \underline{\textbf{77.96}}${}_{\pm 0.47}$           
\\ 
\bottomrule
\end{tabular}
\label{tab:ablation_architecture}
\vspace{-6pt}
\end{table}
\endgroup

\paragraph{Proximal Losses} We utilize KL-divergence as the proximal loss in our primary experiments. However, FedSOL is compatible with various other different proximal objectives. \Autoref{tab:sol_effect} demonstrates the impact of integrating FedSOL with other proximal objectives: FedProx\,\citep{fedprox}, FedNova\,\citep{fednova}, Scaffold\,\citep{scaffold}, FedDyn\,\citep{feddyn}, and Moon\,\citep{moon}. These methods are compared in two distinct scenarios: as an auxiliary objective alongside the original local objective following \autoref{eq_proximal_restriction} (Base), and as proximal perturbation within FedSOL (Combined). The perturbation is applied to the entire model, not just the head, to assess the overall effect of each proximal objective within FedSOL. The results show enhanced performance when combined with FedSOL.

\begingroup
\setlength{\tabcolsep}{2.2pt} 
\renewcommand{\arraystretch}{1.09}
\begin{table}[ht!]
\small
\caption{Comparison of proximal methods when combined with FedSOL ($\rho$=2.0). The heterogeneity is set as LDA ($\alpha$ = 0.1).}
\vspace{-5pt}
\centering
\begin{tabular}{lcccccc} 
\toprule
\multicolumn{1}{c}{\multirow{2}{*}{\textbf{Method}}}
& \multicolumn{2}{c}{\textbf{CIFAR-10}} 
& \multicolumn{2}{c}{\textbf{SVHN}} 
& \multicolumn{2}{c}{\textbf{CINIC-10}}  \\
\multicolumn{1}{c}{}                                 
& Base & Combined        
& Base & Combined        
& Base & Combined              \\ 
\hline\hline
FedProx                         
& 59.80      & \textbf{63.93} \textcolor{green!50!black}{$\uparrow$}                    
& 72.40      & \textbf{84.32} \textcolor{green!50!black}{$\uparrow$}                  
& 40.09      & \textbf{55.25} \textcolor{green!50!black}{$\uparrow$}                    
\\
FedNova    
& 10.00      & \textbf{31.77} \textcolor{green!50!black}{$\uparrow$}                    
& 53.07      & \textbf{79.95} \textcolor{green!50!black}{$\uparrow$}            
& 21.89      & \textbf{42.37} \textcolor{green!50!black}{$\uparrow$}      
\\
Scaffold                                          
& 10.00      & \textbf{62.70} \textcolor{green!50!black}{$\uparrow$}                    
& 21.46      & \textbf{77.52} \textcolor{green!50!black}{$\uparrow$}                  
& 16.89      & \textbf{49.96} \textcolor{green!50!black}{$\uparrow$}                 \\
FedDyn                                          
& 60.80      & \textbf{62.85} \textcolor{green!50!black}{$\uparrow$}                    
& 78.15      & \textbf{79.43} \textcolor{green!50!black}{$\uparrow$}                  
& 48.25     & \textbf{52.17} \textcolor{green!50!black}{$\uparrow$}      
\\
MOON                                                  
& 55.72      & \textbf{60.91} \textcolor{green!50!black}{$\uparrow$}                    
& 29.67      & \textbf{76.82} \textcolor{green!50!black}{$\uparrow$}                  
& 38.15      & \textbf{49.14} \textcolor{green!50!black}{$\uparrow$}                    
\\
\bottomrule
\end{tabular}
\label{tab:sol_effect}
\vspace{-8pt}
\end{table}
\endgroup

\subsection{Ablation Study}
\label{sec4:ablation_study}
\vspace{-2pt}

\begin{figure}[ht!]
    \vspace{-2pt}
    \centering
    \includegraphics[width=0.46\textwidth]{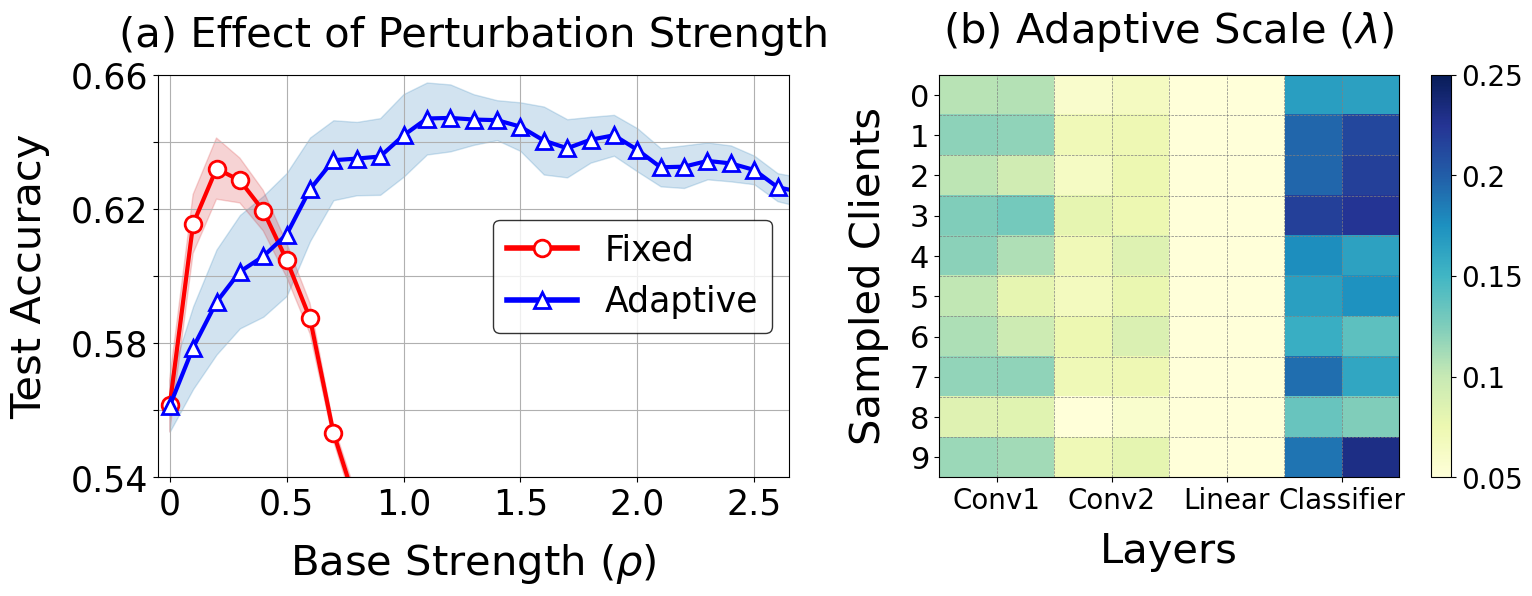}
    \vspace{-4pt}
    \caption{Effect of adaptive perturbation strength in CIFAR-10 ($\alpha$=0.1). (a) Server test accuracy after 300 rounds. (b) Layer-wisely averaged $\lambda$ values of FedSOL ($\rho=1.0$) at round 200.}
    \label{fig:adaptive_strength}
\end{figure}

\paragraph{Adaptive Perturbation Strength}
The effect of the adaptive perturbation is depicted in \autoref{fig:adaptive_strength}. As shown in \autoref{fig:adaptive_strength}\textcolor{red}{(a)}, adaptive perturbation not only improve performances but also reduces sensitivity to the selection of $\rho$. Meanwhile, \autoref{fig:adaptive_strength}\textcolor{red}{(b)} displays the average values for the layer-wise scaling factor $\lambda$ across the local models. The result highlights the increased deviation in the later layers, as a consequence of the data heterogeneity.
\vspace{-7pt}

\paragraph{Partial Perturbation}
The results in \autoref{tab:partial_perturb} reveal that perturbing only the last classifier layer (\textit{Head} in \autoref{tab:partial_perturb}) is sufficient for FedSOL. The performance reaches as high as the full-model perturbation, yet the required computation is considerably lower. Interestingly, perturbing all layers except the head (\textit{Body} in \autoref{tab:partial_perturb}) incurs nearly the same computational cost yet results in diminished performance, highlighting the importance of the later layers.  We conduct further experiments on larger models in \autoref{appendix_partial_perturbation_larger} and discuss the local computational cost in \autoref{appendix_computational_cost}.
\vspace{-2pt}

\vspace{-5pt}
\begingroup
\setlength{\tabcolsep}{2.8pt} 
\renewcommand{\arraystretch}{1.135}
\begin{table}[ht!]
\small
\caption{Effect of partial perturbation in FedSOL on CIFAR10 ($\alpha$=0.1). The FLOPs shows relative computation w.r.t. FedAvg. $\delta$ stands for the computation for the proximal loss.}
\vspace{-5pt}
\centering
\begin{tabular}{lcccccc} 
\toprule
\multicolumn{1}{c}{\multirow{2}{*}{\textbf{Target Position}}} & \multicolumn{5}{c}{\textbf{Perturbation} ($\boldsymbol{\rho}$)}                         & \multirow{2}{*}{\textbf{FLOPs}}  \\
                                        & 0.0                      & 0.5   & 1.0   & 1.5   & 2.0   &                                   \\ 
\hline\hline
All (\textit{full})         & \multirow{3}{*}{56.13} 
& 61.17 & \textbf{64.16} & \textbf{64.38} & \textbf{63.94} & 2$\times$ $+\delta$    \\
Body (\textit{partial})  &                        
& 60.98 & 62.95 & 63.94 & 63.80 & 1.96$\times$ $+\delta$\\
Head (\textit{partial})   &                        
& \textbf{62.65} & 63.62 & 64.13 & 63.25 & 1.33$\times$ $+\delta$ \\
\bottomrule
\end{tabular}
\label{tab:partial_perturb}
\vspace{-5.5pt}
\end{table}
\endgroup

%% file: materials/tables/main_exp.tex
\begingroup
\setlength{\tabcolsep}{0.935pt} 
\renewcommand{\arraystretch}{1.15}
\begin{table*}[ht!]
\caption{
 Test accuracy@1(\%) comparison among baselines and FedSoL on different datasets. The values in the parenthesis are the standard deviation. The arrow (\scriptsize{\textcolor{red}{$\downarrow$}},\, \scriptsize{\textcolor{green!50!black}{$\uparrow$}}\normalsize)\, shows the comparison to the FedAvg. We set $s \in \{ 2, 3, 5, 10 \}$ and $\alpha \in \{0.05, 0.1, 0.3, 0.5\}$ for CIFAR-10 datasets, whereas $s=2$ and $\alpha = 0.1$ for the others.}
 \vspace{-5pt}
\centering
\small
\begin{tabular}{lrrrrrrrrr}
\toprule
\multicolumn{10}{c}{\normalsize{\textbf{Non-IID Partition Strategy : Sharding}}}
\small
\\ 
\hline
\multicolumn{1}{c}{\multirow{2}{*}{\textbf{Method}}} 
& \multicolumn{1}{c}{\multirow{2}{*}{\textbf{MNIST}}} 
& \multicolumn{4}{c}{\textbf{CIFAR-10}}                                             
& \multicolumn{1}{c}{\multirow{2}{*}{\textbf{SVHN}}} 
& \multicolumn{1}{c}{\multirow{2}{*}{\textbf{CINIC-10}}} 
& \multicolumn{1}{c}{\multirow{2}{*}{\textbf{PathMNIST}}} 
& \multicolumn{1}{c}{\multirow{2}{*}{\textbf{TissueMNIST}}}  
\\
\multicolumn{1}{c}{}                                 
& \multicolumn{1}{c}{}                                 
& \multicolumn{1}{c}{$s$ = 2}    
& \multicolumn{1}{c}{$s$ = 3}    
& \multicolumn{1}{c}{$s$ = 5}   
& \multicolumn{1}{c}{$s$ = 10}  
& \multicolumn{1}{c}{}                                
& \multicolumn{1}{c}{}                                    
& \multicolumn{1}{c}{}                                     
& \multicolumn{1}{c}{}                                        
\\ 
\hline\hline

FedAvg\,\citep{fedavg}
& 96.16${}_{(0.19)}\,\,\,$ 
& 51.48${}_{(3.41)}\,\,\,$ 
& 62.94${}_{(0.00)}\,\,\,$ 
& 70.96${}_{(0.91)}\,\,\,$ 
& 74.60${}_{(0.88)}\,\,\,$ 
& 73.63${}_{(3.16)}\,\,\,$ 
& 42.40${}_{(2.70)}\,\,\,$ 
& 57.40${}_{(1.48)}\,\,\,$
& 49.36${}_{(1.64)}\,\,\,$ 
\\
\hline

FedProx\,\citep{fedprox}
& 95.86${}_{(0.12)}\,$\scriptsize{\textcolor{red}{$\downarrow$}} 
& 52.80${}_{(2.66)}\,$\scriptsize{\textcolor{green!50!black}{$\uparrow$}} 
& 58.19${}_{(0.55)}\,$\scriptsize{\textcolor{red}{$\downarrow$}} 
& 64.71${}_{(0.74)}\,$\scriptsize{\textcolor{red}{$\downarrow$}} 
& 69.37${}_{(1.21)}\,$\scriptsize{\textcolor{red}{$\downarrow$}} 
& 71.09${}_{(3.13)}\,$\scriptsize{\textcolor{red}{$\downarrow$}} 
& 40.00${}_{(3.01)}\,$\scriptsize{\textcolor{red}{$\downarrow$}} 
& 60.77${}_{(3.64)}\,$\scriptsize{\textcolor{green!50!black}{$\uparrow$}}
& 48.20${}_{(1.95)}\,$\scriptsize{\textcolor{red}{$\downarrow$}} \\

FedNova\,\citep{fednova}
& 94.13${}_{(0.36)}\,$\scriptsize{\textcolor{red}{$\downarrow$}} 
& 46.89${}_{(2.57)}\,$\scriptsize{\textcolor{red}{$\downarrow$}} 
& 61.12${}_{(0.88)}\,$\scriptsize{\textcolor{red}{$\downarrow$}} 
& 67.11${}_{(0.25)}\,$\scriptsize{\textcolor{red}{$\downarrow$}} 
& 70.59${}_{(0.52)}\,$\scriptsize{\textcolor{red}{$\downarrow$}} 
& 67.35${}_{(2.84)}\,$\scriptsize{\textcolor{red}{$\downarrow$}} 
& 40.94${}_{(2.29)}\,$\scriptsize{\textcolor{red}{$\downarrow$}} 
& 58.85${}_{(4.10)}\,$\scriptsize{\textcolor{green!50!black}{$\uparrow$}}
& 36.44${}_{(0.95)}\,$\scriptsize{\textcolor{red}{$\downarrow$}} \\

Scaffold\,\citep{scaffold}
& 95.91${}_{(0.18)}\,$\scriptsize{\textcolor{red}{$\downarrow$}} 
& 62.60${}_{(0.70)}\,$\scriptsize{\textcolor{green!50!black}{$\uparrow$}}
& 68.53${}_{(0.99)}\,$\scriptsize{\textcolor{green!50!black}{$\uparrow$}}
& 74.28${}_{(0.39)}\,$\scriptsize{\textcolor{green!50!black}{$\uparrow$}}
& 76.71${}_{(0.16)}\,$\scriptsize{\textcolor{green!50!black}{$\uparrow$}}
& 77.84${}_{(2.28)}\,$\scriptsize{\textcolor{green!50!black}{$\uparrow$}}
& 47.76${}_{(0.45)}\,$\scriptsize{\textcolor{green!50!black}{$\uparrow$}}
& 71.12${}_{(1.04)}\,$\scriptsize{\textcolor{green!50!black}{$\uparrow$}}
& 30.99${}_{(6.09)}\,$\scriptsize{\textcolor{red}{$\downarrow$}} \\


FedNTD\,\citep{fedntd}
& 96.62${}_{(0.06)}\,$\scriptsize{\textcolor{green!50!black}{$\uparrow$}}
& \textbf{\underline{67.25}}${}_{(1.08)}\,$\scriptsize{\textcolor{green!50!black}{$\uparrow$}} 
& \textbf{\underline{70.47}}${}_{(0.33)}\,$\scriptsize{\textcolor{green!50!black}{$\uparrow$}} 
& 75.21${}_{(0.39)}\,$\scriptsize{\textcolor{green!50!black}{$\uparrow$}} 
& 76.46${}_{(0.07)}\,$\scriptsize{\textcolor{green!50!black}{$\uparrow$}} 
& \textbf{\underline{85.30}}${}_{(0.78)}\,$\scriptsize{\textcolor{green!50!black}{$\uparrow$}}
& 52.72${}_{(1.12)}\,$\scriptsize{\textcolor{green!50!black}{$\uparrow$}}
& 65.00${}_{(1.26)}\,$\scriptsize{\textcolor{green!50!black}{$\uparrow$}}
& 52.63${}_{(0.59)}\,$\scriptsize{\textcolor{green!50!black}{$\uparrow$}} \\

FedSAM\,\citep{fedsam}
& 96.12${}_{(0.19)}\,$\scriptsize{\textcolor{red}{$\downarrow$}} 
& 51.85${}_{(3.14)}\,$\scriptsize{\textcolor{green!50!black}{$\uparrow$}} 
& 60.90${}_{(0.93)}\,$\scriptsize{\textcolor{red}{$\downarrow$}} 
& 69.29${}_{(0.39)}\,$\scriptsize{\textcolor{red}{$\downarrow$}} 
& 72.98${}_{(0.34)}\,$\scriptsize{\textcolor{red}{$\downarrow$}} 
& 65.85${}_{(3.77)}\,$\scriptsize{\textcolor{red}{$\downarrow$}} 
& 45.91${}_{(2.02)}\,$\scriptsize{\textcolor{green!50!black}{$\uparrow$}} 
& 67.32${}_{(3.15)}\,$\scriptsize{\textcolor{green!50!black}{$\uparrow$}}  
& 49.62${}_{(1.61)}\,$\scriptsize{\textcolor{green!50!black}{$\uparrow$}}  \\

FedASAM\,\citep{fedasam}
&  \textbf{\underline{97.08}}${}_{(0.15)}\,$\scriptsize{\textcolor{green!50!black}{$\uparrow$}} 
& 52.08${}_{(2.19)}\,$\scriptsize{\textcolor{green!50!black}{$\uparrow$}}
& 63.24${}_{(1.16)}\,$\scriptsize{\textcolor{green!50!black}{$\uparrow$}} 
& 70.95${}_{(0.76)}\,$\scriptsize{\textcolor{red}{$\downarrow$}} 
& 74.74${}_{(0.88)}\,$\scriptsize{\textcolor{green!50!black}{$\uparrow$}} 
& 79.48${}_{(2.17)}\,$\scriptsize{\textcolor{red}{$\downarrow$}} 
& 43.15${}_{(2.73)}\,$\scriptsize{\textcolor{green!50!black}{$\uparrow$}}
& 59.47${}_{(2.91)}\,$\scriptsize{\textcolor{green!50!black}{$\uparrow$}}
& 49.46${}_{(1.91)}\,$\scriptsize{\textcolor{green!50!black}{$\uparrow$}} \\

\hline
\textbf{FedSOL \scriptsize{(Ours)}}  
& \textbf{\underline{97.15}}${}_{(0.08)}\,$\scriptsize{\textcolor{green!50!black}{$\uparrow$}}
& \textbf{\underline{66.72}}${}_{(0.61)}\,$\scriptsize{\textcolor{green!50!black}{$\uparrow$}}
& \textbf{\underline{69.88}}${}_{(0.15)}\,$\scriptsize{\textcolor{green!50!black}{$\uparrow$}}
& \textbf{\underline{75.82}}${}_{(0.34)}\,$\scriptsize{\textcolor{green!50!black}{$\uparrow$}}
& \textbf{\underline{77.79}}${}_{(0.19)}\,$\scriptsize{\textcolor{green!50!black}{$\uparrow$}}
& \textbf{\underline{85.18}}${}_{(0.37)}\,$\scriptsize{\textcolor{green!50!black}{$\uparrow$}}
& \textbf{\underline{55.17}}${}_{(0.32)}\,$\scriptsize{\textcolor{green!50!black}{$\uparrow$}}
& \textbf{\underline{73.85}}${}_{(1.55)}\,$\scriptsize{\textcolor{green!50!black}{$\uparrow$}}
& \textbf{\underline{53.42}}${}_{(0.46)}\,$\scriptsize{\textcolor{green!50!black}{$\uparrow$}}
\\

\bottomrule\toprule
\addlinespace[0.25em]
\multicolumn{10}{c}{\normalsize{\textbf{Non-IID Partition Strategy : LDA}}}
\small
\\ 
\hline
\multicolumn{1}{c}{\multirow{2}{*}{\textbf{Method}}}  
& \multicolumn{1}{c}{\multirow{2}{*}{\textbf{MNIST}}} 
& \multicolumn{4}{c}{\textbf{CIFAR-10}}                                             
& \multicolumn{1}{c}{\multirow{2}{*}{\textbf{SVHN}}} 
& \multicolumn{1}{c}{\multirow{2}{*}{\textbf{CINIC-10}}} 
& \multicolumn{1}{c}{\multirow{2}{*}{\textbf{PathMNIST}}} 
& \multicolumn{1}{c}{\multirow{2}{*}{\textbf{TissueMNIST}}}  
\\
\multicolumn{1}{c}{}                                  
& \multicolumn{1}{c}{}                                 
& \multicolumn{1}{c}{$\alpha$ = 0.05} 
& \multicolumn{1}{c}{$\alpha$ = 0.1} 
& \multicolumn{1}{c}{$\alpha$ = 0.3} 
& \multicolumn{1}{c}{$\alpha$ = 0.5} 
& \multicolumn{1}{c}{}                              
& \multicolumn{1}{c}{}                                    
& \multicolumn{1}{c}{}                                    
& \multicolumn{1}{c}{}                                        
\\
\hline\hline
FedAvg\,\citep{fedavg}
& 96.11${}_{(0.19)}\,\,\,$ 
& 42.27${}_{(1.34)}\,\,\,$ 
& 56.13${}_{(0.78)}\,\,\,$ 
& 67.32${}_{(0.94)}\,\,\,$ 
& 73.90${}_{(0.66)}\,\,\,$ 
& 55.36${}_{(4.85)}\,\,\,$ 
& 36.49${}_{(4.37)}\,\,\,$ 
& 65.98${}_{(4.76)}\,\,\,$
& 42.78${}_{(2.03)}\,\,\,$ 
\\
\hline

FedProx\,\citep{fedprox}
& 96.05${}_{(0.13)}\,$\scriptsize{\textcolor{red}{$\downarrow$}} 
& 50.58${}_{(0.57)}\,$\scriptsize{\textcolor{green!50!black}{$\uparrow$}} 
& 59.80${}_{(1.12)}\,$\scriptsize{\textcolor{green!50!black}{$\uparrow$}} 
& 68.39${}_{(0.81)}\,$\scriptsize{\textcolor{green!50!black}{$\uparrow$}} 
& 72.87${}_{(0.55)}\,$\scriptsize{\textcolor{green!50!black}{$\uparrow$}} 
& 72.40${}_{(3.15)}\,$\scriptsize{\textcolor{green!50!black}{$\uparrow$}} 
& 40.09${}_{(3.97)}\,$\scriptsize{\textcolor{green!50!black}{$\uparrow$}} 
& 70.44${}_{(1.92)}\,$\scriptsize{\textcolor{green!50!black}{$\uparrow$}} 
& 52.25${}_{(1.40)}\,$\scriptsize{\textcolor{green!50!black}{$\uparrow$}}  \\

FedNova\,\citep{fednova}
& 88.24${}_{(1.37)}\,$\scriptsize{\textcolor{red}{$\downarrow$}} 
& 10.00${}_{(\scriptsize{\textit{Failed}})}\,$\scriptsize{\textcolor{red}{$\downarrow$}} 
& 10.00${}_{(\scriptsize{\textit{Failed}})}\,$\scriptsize{\textcolor{red}{$\downarrow$}} 
& 64.67${}_{(0.77)}\,$\scriptsize{\textcolor{red}{$\downarrow$}} 
& 70.04${}_{(0.45)}\,$\scriptsize{\textcolor{red}{$\downarrow$}} 
& 53.07${}_{(3.30)}\,$\scriptsize{\textcolor{red}{$\downarrow$}} 
& 21.89${}_{(1.71)}\,$\scriptsize{\textcolor{red}{$\downarrow$}} 
& 38.94${}_{(2.34)}\,$\scriptsize{\textcolor{red}{$\downarrow$}} 
& 15.03${}_{(3.74)}\,$\scriptsize{\textcolor{red}{$\downarrow$}} \\

Scaffold\,\citep{scaffold}
& 94.18${}_{(0.32)}\,$\scriptsize{\textcolor{red}{$\downarrow$}} 
& 10.00${}_{(\scriptsize{\textit{Failed}})}\,$\scriptsize{\textcolor{red}{$\downarrow$}} 
& 10.00${}_{(\scriptsize{\textit{Failed}})}\,$\scriptsize{\textcolor{red}{$\downarrow$}} 
& \underline{\textbf{71.92}}${}_{(0.17)}\,$\scriptsize{\textcolor{green!50!black}{$\uparrow$}} 
& \textbf{\underline{75.49}}${}_{(0.21)}\,$\scriptsize{\textcolor{green!50!black}{$\uparrow$}}  
& 21.46${}_{(1.75)}\,$\scriptsize{\textcolor{red}{$\downarrow$}} 
& 16.89${}_{(2.25)}\,$\scriptsize{\textcolor{red}{$\downarrow$}} 
& 18.07${}_{(0.04)}\,$\scriptsize{\textcolor{red}{$\downarrow$}} 
& 32.04${}_{(0.07)}\,$\scriptsize{\textcolor{red}{$\downarrow$}} \\


FedNTD\,\citep{fedntd}
& 96.97${}_{(0.27)}\,$\scriptsize{\textcolor{green!50!black}{$\uparrow$}}
& 58.08${}_{(0.48)}\,$\scriptsize{\textcolor{green!50!black}{$\uparrow$}} 
& \underline{\textbf{63.16}}${}_{(1.02)}\,$\scriptsize{\textcolor{green!50!black}{$\uparrow$}}
& 71.56${}_{(0.26)}\,$\scriptsize{\textcolor{green!50!black}{$\uparrow$}}
& 74.91${}_{(0.33)}\,$\scriptsize{\textcolor{green!50!black}{$\uparrow$}}
& 79.25${}_{(0.61)}\,$\scriptsize{\textcolor{green!50!black}{$\uparrow$}}
& 50.22${}_{(3.71)}\,$\scriptsize{\textcolor{green!50!black}{$\uparrow$}}
& 74.26${}_{(1.25)}\,$\scriptsize{\textcolor{green!50!black}{$\uparrow$}}
& 44.55${}_{(1.95)}\,$\scriptsize{\textcolor{green!50!black}{$\uparrow$}} \\

FedSAM\,\citep{fedsam}
& 95.72${}_{(0.43)}\,$\scriptsize{\textcolor{red}{$\downarrow$}} 
& 36.14${}_{(1.21)}\,$\scriptsize{\textcolor{red}{$\downarrow$}} 
& 52.14${}_{(0.94)}\,$\scriptsize{\textcolor{red}{$\downarrow$}} 
& 64.83${}_{(0.56)}\,$\scriptsize{\textcolor{red}{$\downarrow$}} 
& 70.74${}_{(0.40)}\,$\scriptsize{\textcolor{red}{$\downarrow$}} 
& 13.27${}_{(2.78)}\,$\scriptsize{\textcolor{red}{$\downarrow$}} 
& 36.70${}_{(4.28)}\,$\scriptsize{\textcolor{green!50!black}{$\uparrow$}} 
& 66.64${}_{(3.76)}\,$\scriptsize{\textcolor{green!50!black}{$\uparrow$}}
& 44.07${}_{(3.02)}\,$\scriptsize{\textcolor{green!50!black}{$\uparrow$}} \\

FedASAM\,\citep{fedasam}
& 96.60${}_{(0.10)}\,$\scriptsize{\textcolor{green!50!black}{$\uparrow$}}  
& 43.12${}_{(1.25)}\,$\scriptsize{\textcolor{green!50!black}{$\uparrow$}}  
& 57.00${}_{(0.30)}\,$\scriptsize{\textcolor{green!50!black}{$\uparrow$}}  
& 67.45${}_{(0.92)}\,$\scriptsize{\textcolor{green!50!black}{$\uparrow$}}  
& 73.91${}_{(0.51)}\,$\scriptsize{\textcolor{green!50!black}{$\uparrow$}}  
& 60.25${}_{(4.56)}\,$\scriptsize{\textcolor{green!50!black}{$\uparrow$}}  
& 36.93${}_{(4.60)}\,$\scriptsize{\textcolor{green!50!black}{$\uparrow$}}
& 69.45${}_{(3.19)}\,$\scriptsize{\textcolor{green!50!black}{$\uparrow$}}
& 42.73${}_{(2.35)}\,$\scriptsize{\textcolor{green!50!black}{$\uparrow$}} \\

\hline
\textbf{FedSOL \scriptsize{(Ours)}}  
& \textbf{\underline{97.44}}${}_{(0.11)}\,$\scriptsize{\textcolor{green!50!black}{$\uparrow$}}
& \textbf{\underline{60.01}}${}_{(0.30)}\,$\scriptsize{\textcolor{green!50!black}{$\uparrow$}}
& \textbf{\underline{64.13}}${}_{(0.46)}\,$\scriptsize{\textcolor{green!50!black}{$\uparrow$}}
& \textbf{\underline{71.94}}${}_{(0.57)}\,$\scriptsize{\textcolor{green!50!black}{$\uparrow$}}
& \textbf{\underline{75.60}}${}_{(0.32)}\,$\scriptsize{\textcolor{green!50!black}{$\uparrow$}}
& \textbf{\underline{83.92}}${}_{(0.29)}\,$\scriptsize{\textcolor{green!50!black}{$\uparrow$}}
& \textbf{\underline{55.07}}${}_{(1.48)}\,$\scriptsize{\textcolor{green!50!black}{$\uparrow$}}
& \textbf{\underline{78.88}}${}_{(0.46)}\,$\scriptsize{\textcolor{green!50!black}{$\uparrow$}}
& \textbf{\underline{53.40}}${}_{(0.85)}\,$\scriptsize{\textcolor{green!50!black}{$\uparrow$}}
\\
\bottomrule
\end{tabular}
\label{tab:main_std}
\vspace{-5pt}
\end{table*}
\endgroup

%% file: sec/05-analysis.tex
\section{Analysis}
\label{sec:analysis}
\begin{figure}[ht!]
    \centering
    \includegraphics[width=0.465\textwidth]{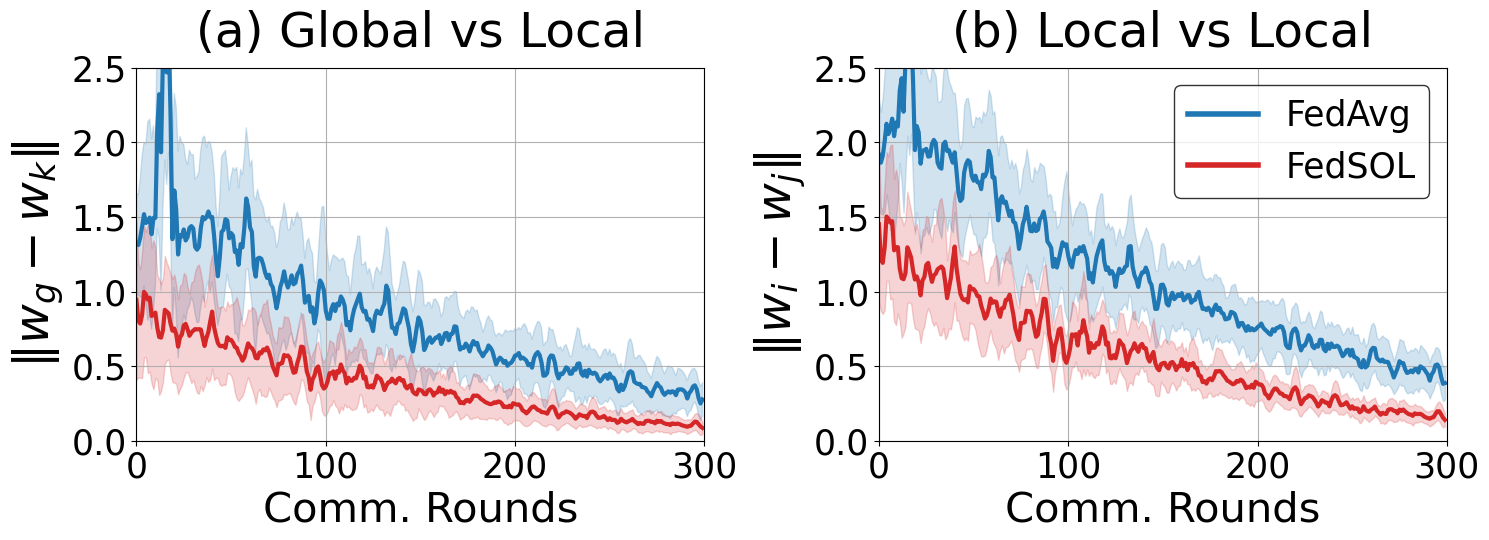}
    \vspace{-3pt}
    \caption{Analysis on weight divergence in FedAvg and FedSOL ($\rho$=2.0) on CIFAR-10 LDA ($\alpha$=0.1). (a) shows global-local model divergence, while (b) presents the divergence across local models.}
    \label{fig:weight_divergence}
\end{figure}

\vspace{-1pt}
\paragraph{Weight Divergence}
To assess the deviation of local learning from the global model, we measure the L2 distance between models: $\lVert {w}_{g} - {w}_{k} \rVert$, where $w_g$ is the global model and $w_k$ is the client $k$'s trained local model. \autoref{fig:weight_divergence} shows the results averaged across sampled clients. In \autoref{fig:weight_divergence}\textcolor{red}{(a)}, FedSOL effectively reduces the divergence, ensuring that local models stay closely aligned with the global model. In \autoref{fig:weight_divergence}\textcolor{red}{(b)}, FedSOL also promotes increased consistency among local models, reducing  their mutual divergence.
\vspace{-1pt}

\begin{figure}[ht!]
    \centering
\includegraphics[width=0.47\textwidth]{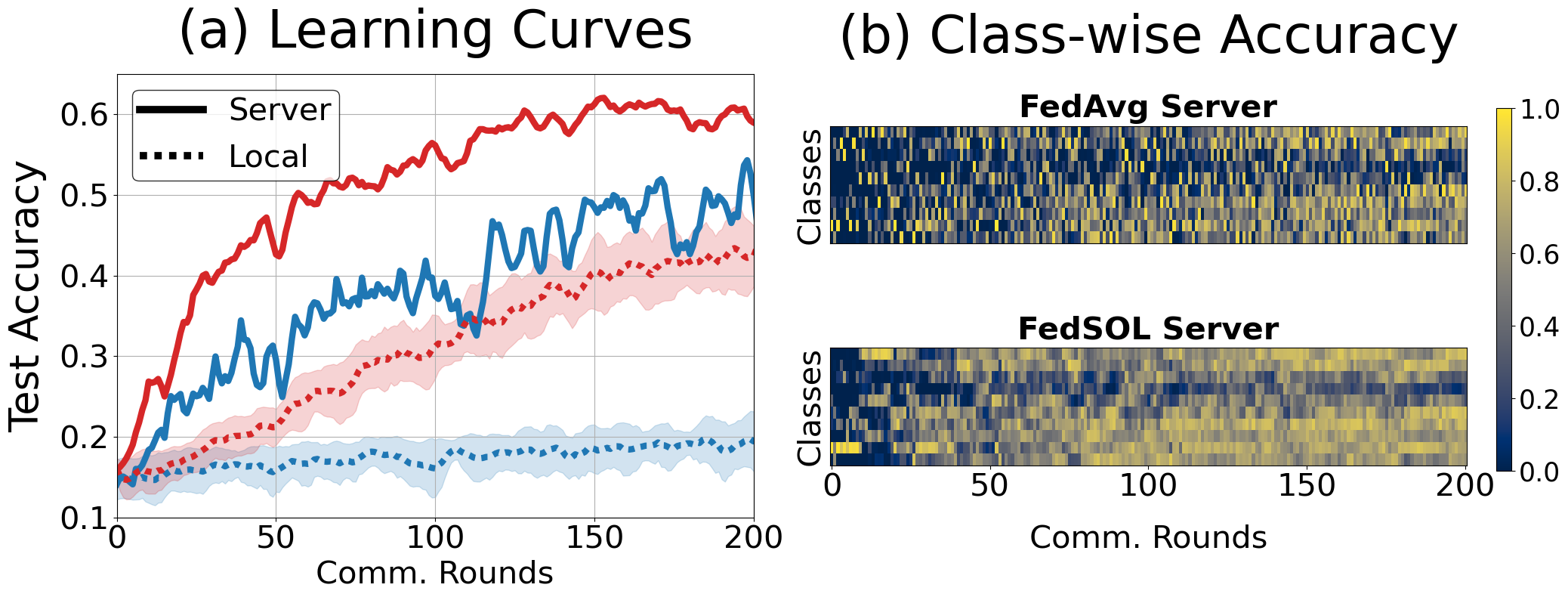}
\vspace{-4pt}
\caption{Comparison of FedAvg (\textcolor{blue}{\textit{blue}} lines) and FedSOL ($\rho$=2.0) (\textcolor{red}{\textit{red}} lines) on CIFAR-10 (s=2). (a) learning curves for global and local models. The shaded areas reflect standard deviation across clients. (b) class-wise accuracy of the global model.}
\label{fig:knowledge_preservation}
\end{figure}

\paragraph{Knowledge Preservation}
We examine the effect of FedSOL on knowledge preservation during local learning. The results in \autoref{fig:knowledge_preservation} show performance on the global distribution. As depicted in \autoref{fig:knowledge_preservation},  local models using FedAvg experience a significant drop in performance on the global distribution after local learning. In contrast, FedSOL maintains high performance on the global distribution, indicating that its orthogonal learning approach effectively preserves global knowledge. Further analysis of class-wise accuracy for FedAvg and FedSOL server models is presented in \autoref{fig:knowledge_preservation}\textcolor{red}{(b)}. The results demonstrate that while FedAvg exhibits significant fluctuations and inconsistent class-wise accuracy, FedSOL consistently maintains its class-wise accuracy as communication proceeds.

\vspace{-2pt}

\vspace{-14pt}
\paragraph{Smoothness of Loss Landscape}

\begin{figure}[ht!]
    \centering
    \includegraphics[width=0.465\textwidth]{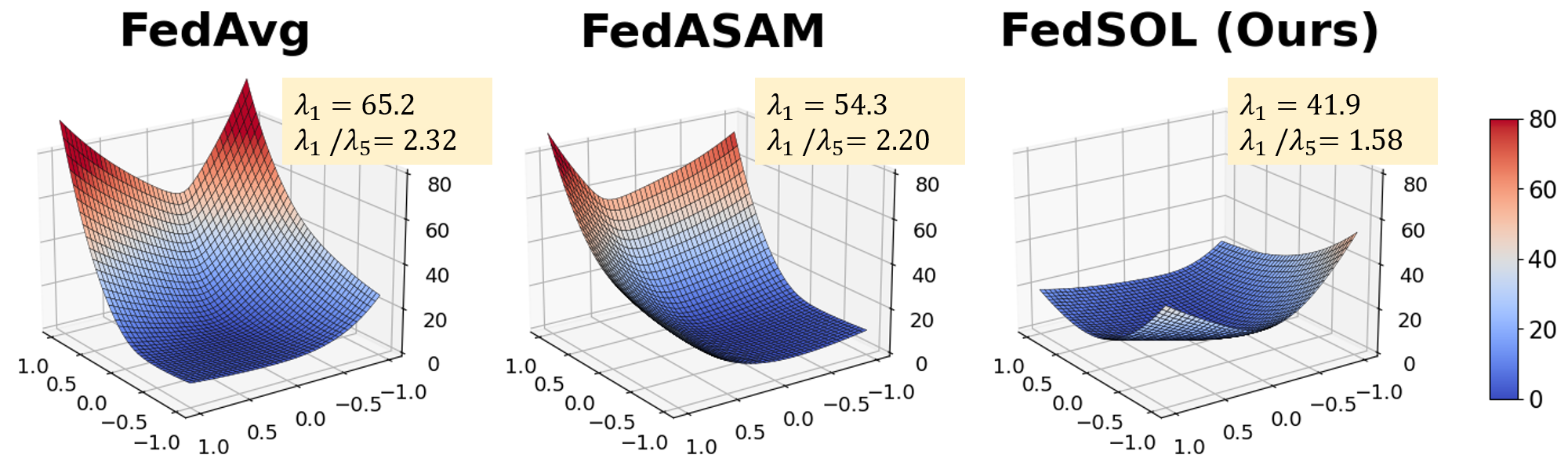}
    \vspace{-2pt}
    \caption{Loss landscape visualization of global model on CIFAR-10 LDA ($\alpha$=0.1). The ${\lambda}_1$ and ${\lambda}_5$ in each figure stand for the top-1 and top-5 eigenvalues of the Hessian matrix.}
    \label{fig:loss_landscape}
\end{figure}

It has been suggested that models at flatter minima more easily preserve previous knowledge after adapting to new distributions\,\citep{fs_dgpm, lipschitz_continual, cpr, dfgp}. In \autoref{fig:loss_landscape}, we visualize the loss landscapes\,\citep{visualizing_loss_landscape} of global models obtained from FedAvg, FedASAM, and FedSOL. In these plots, each axis corresponds to one of the two dominant eigenvectors (top-1 and top-2) of the Hessian matrix, representing the directions of the most significant shifts in the loss landscape. Along with each landscape, we include the value of the dominant eigenvalue (${\lambda}_1$) and its ratio to the fifth-largest eigenvalue (${\lambda}_1 / {\lambda}_5$), following the criteria used in \citep{sam, ssam}. The smaller ratio observed in FedSOL indicates that variations in the loss are more evenly distributed across different directions. These results demonstrate FedSOL's effectiveness in smoothing the loss landscape.

%% file: sec/06-related-work.tex
\section{Related Work}
\vspace{-2pt}

\paragraph{Federated Learning\,(FL)}
Federated learning is a distributed learning paradigm to train models without directly accessing private client data\,\citep{federated_learning, federated_optimization}. The standard algorithm, FedAvg\,\citep{fedavg}, aggregates trained local models by averaging their parameters. FedAvg ideally performs well when all client devices are active and IID distributed\,\citep{fast_local_sgd, minibatch_sgd}. While various FL algorithms have been introduced, they commonly conduct parameter averaging in a certain manner\,\citep{fedprox, scaffold, fedntd, fedftg}. However, its performance significantly degrades under heterogeneous data distributions among clients\,\citep{fl_with_noniid, On_the_convergence_fedavg_noniid, advances_open_problems_fl}. Our work aims to address this data heterogeneity issue by modifying the local learning.

\vspace{-5.5pt}
\paragraph{Proximal Restriction in FL} A prevalent strategy to address data heterogeneity in FL involves introducing a proximal objective into local learning as an auxiliary loss\,\citep{fedntd, fedprox, scaffold, moon}. This approach aims to restrict the deviation of local learning induced by the biased local distributions. For example, FedProx\,\citep{fedprox} employs an $L_2$ distance between models, while Scaffold\,\citep{scaffold} employs the estimated global direction as a control variate to adjust local gradients. However, such proximal objectives may hinder the acquisition of new knowledge during local learning due to conflicts with the local objective\,\citep{fedsam, fedalign}. In our work, we leverage the proximal objective for weight perturbation, thereby enabling local learning to be orthogonal to the proximal objective.

\vspace{-5.5pt}
\paragraph{Orthogonal Learning in CL} In CL, many approaches have adopted orthogonal learning, which align new task gradients orthogonal to old task loss spaces\,\citep{fs_dgpm, gpm, ogd, gem}. This typically involves retaining data or gradients as memory\,\citep{agem, cl_low_rank}. In FL, few recent attempts have applied similar strategies\,\citep{fot, gradma}. For example, FOT\,\citep{fot} uses a random Gaussian matrix with SVD, and GradMA\,\citep{gradma} employs gradient memory and quadratic programming, both requiring substantial computational resources. In our work, we use the proximal objective to preserve knowledge and update using a local gradient orthogonal to it. However, we observe that directly opposing the proximal gradient  can adversely affect performance, a finding that aligns with recent CL research about the drawbacks of direct gradient projection\,\citep{cl_rethinking_gradient_projection, dfgp}. To address this, we propose FedSOL, which implicitly finds the orthogonal local gradient.

%% file: sec/07-conclusion.tex
\vspace{-2pt}
\section{Conclusion}
\vspace{-2pt}
In this study, we propose a novel FL algorithm, FedSOL. Inspired by CL, FedSOL identifies the local gradient that is orthogonal to the proximal gradient during local learning. This orthogonal learning strategy helps to maintain previous global knowledge throughout the local learning process. We conduct extensive experiments to validate the efficacy of FedSOL and demonstrate its benefits in FL.

%% file: sec/appendix.tex
\clearpage
\setcounter{page}{1}
\maketitlesupplementary

\appendix
\setlength{\textfloatsep}{15pt}

\section{Table of Notations}
\vspace{-2pt}

\begingroup
\setlength{\tabcolsep}{3.0pt} 
\renewcommand{\arraystretch}{1.1}
\begin{table}[h!]
\vspace{-8pt}
\caption{Table of Notations throughout the paper.}
\vspace{-6pt}
\begin{tabularx}{0.49\textwidth}{p{0.18\textwidth}X}
      \toprule
      {Indices: } \\
      $k$ & Index for clients ($k\in[K]$)\\
      $g$ & Index for global server\\
      \midrule
      {Environment:} \\
      $\mathcal{D}$    & Whole dataset\\
      $\mathcal{D}^k$  & Local dataset of the $k$-th client\\
      $\alpha$   & Concentration parameter for the Dirichlet distribution\\
      $s$        & The number of shards per user\\
      \midrule
      {FL algorithms:} \\
      $\beta, \mu$    & Multiplicative coefficient for the proximal loss\\
      $\gamma$  & Learning rate\\
      $\tau$   & Temperature to be divided in the softmax probability distribution\\
      $\rho$   & Perturbation Strength for SAM-related algorithms\\
      $\boldsymbol{\Lambda}$ & Vector consists with scaling parameters for perturbation vector in SAM-related algorithms\\
      \midrule
      {Weights: } \\

      $\bw_g$   & Weight of the global server model\\
      $\bw_k$ & Weight of the $k$-th client model\\
      $\Vert \bwg - \bwk \Vert$ &  Collection of $L^2$-norm between server and client models, among all rounds.\\
      \midrule
      {Objective Functions: }\\
      $\mathcal{L}^k_{\mathrm{local}}$  & Local objective for the $k$-th client\\
      $\mathcal{L}^k_{p}$  & Proximal objective for the $k$-th client\\
      \bottomrule
\end{tabularx}
\vspace{-8pt}
\end{table}
\endgroup

\section{Experimental Setups}
\vspace{-2pt}
\label{appendix_exeprimental_setups}
The code is implemented by PyTorch\,\citep{pytorch}. The overall code structure is based on FedML\,\citep{fedml} library with some modifications for simplicity. We use 2 A6000 GPU cards, but without Multi-GPU training.

\subsection{Model Architecture} In our primary experiments, we use the model architecture used in FedAvg\,\citep{fedavg}, which consists of two convolutional layers with subsequent max-pooling layers, and two fully-connected layers. The same model architecture is also used in \citep{fedntd, moon, ccvr}. We also conduct further experiments on ResNet-18\,\citep{resnet}, Vgg-11\,\citep{vggnet}, and SL-ViT\,\citep{vit_small_datasets}. For SL-ViT, we resize $28\times28$-sized images into $32\times32$ to accommodate the required minimum size for the patch embedding.

\subsection{Datasets}
\vspace{-2pt}
To validate our approach, we employ 6 distinct datasets, as listed below. The values in the parentheses denote the number of samples used to \textit{train} and \textit{test}, respectively.
\begin{itemize}
    \item \textbf{MNIST} \citep{mnist} (60,000 / 10,000): contains hand-written digits images, ranging from 0 to 9. The data is augmented using Random Cropping, Random Horizontal Flipping, and Normalization. The data is converted to 3-channel RGB images.
    
    \item \textbf{CIFAR-10} \citep{cifar} (50,000 / 10,000):  contains a labeled subset of 80 Million Tiny Images \citep{80million} for 10 different classes. The data is augmented using Random Cropping, Horizontal Flipping, Normalization, and Cutout\,\citep{cutout}.
    
    \item \textbf{SVHN} \citep{svhn} (73,257 / 26,032): contains digits of house numbers obtained from \textit{Google Street View}. The data is augmented using Random Cropping, Random Horizontal Flipping, and Normalization.
    
    \item \textbf{CINIC-10} \citep{cinic10} (90,000 / 90,000): is a combination of CIFAR and downsized ImageNet \citep{imagenet}, which is compiled to serve as a bridge between the two datasets. The data is augmented using Random Cropping, Random Horizontal Flipping, and Normalization.
    
    \item \textbf{PathMNIST} \citep{medmnist} (110,000 / 7,180): contains non-overlapping patches from Hematoxylin \& Eosin stained colorectal histology slide images. The data is augmented using Random Horizontal Flipping, and Normalization.
    
    \item \textbf{TissueMNIST} \citep{medmnist} (189,106 / 47,280): contains microscope images of human kidney cortex cells, which are segmented from 3 reference tissue specimens. The data is augmented using Random Horizontal Flipping, and Normalization. The data is converted to RGB images.
\end{itemize}
\begin{figure}[ht!]
    \centering
    \includegraphics[width=0.47\textwidth]{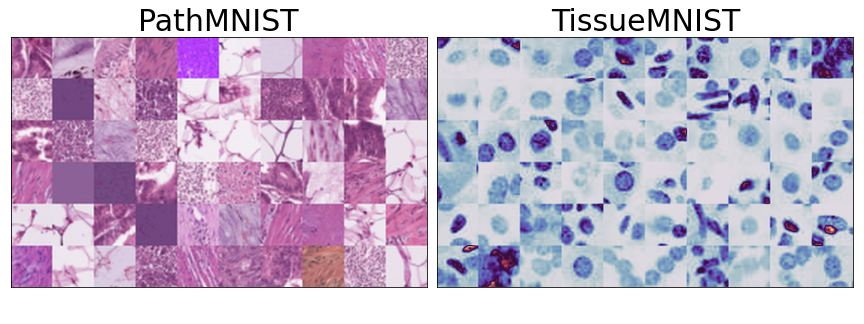}
    \caption{Example images from PathMNIST datasets and TissueMNIST datasets.}
    \label{fig:medmnist}
\end{figure}

\vspace{2pt}
\noindent Note that we evaluate our algorithm is on medical imaging datasets - a crucial practical application of federated learning \citep{fl_survey_technologies_applications, federatedML_concepts_applications}. Illustrative examples of the images are illustrated in \autoref{fig:medmnist}.

\subsection{Non-IID Partition Strategy} To comprehensively address the data heterogeneity issue in federated learning, we distribute the local datasets using the following two distinct data partition strategies: (i) \textbf{Sharding} and (ii) \textbf{Latent Dirichlet Allocation} (LDA). 

\begin{itemize}
    \vspace{1.5pt}
    \item \textbf{(i) Sharding} \citep{fedavg, fedbabu, fedntd}: sorts the data by label and divide the data into shards of the same size, and distribute them to the clients. In this strategy, the heterogeneity level increases as the shard per user, $s$, becomes smaller, and vice versa. As the number of shards is the same across all the clients, \textit{the dataset size is identical for each client}.
    \item \textbf{(ii) Latent Dirichlet Allocation (LDA)} \citep{ccvr, fedma, moon}: allocates the data samples of class $c$ to each client $k$ with the probability $p_c$, where $p_c \approx \text{Dir}(\alpha)$. In this strategy, \textit{both the distribution and dataset size differ for each client}. The heterogeneity level increases as the concentration parameter, $\alpha$, becomes smaller, and vice versa.
\end{itemize}
Note that although only the statistical distributions varies across the clients in Sharding strategy, both the distribution and dataset size differ in LDA strategy.

\subsection{Learning Setups} We use a momentum SGD optimizer with an initial learning rate of 0.01, a momentum value of 0.9, and weight decay 1e-5. The momentum is employed only for local learning and is not uploaded to the server. Note that SAM optimization also requires its base optimizer, which performs the parameter update using the obtained gradient at the perturbed weights. The learning rate is decays with a factor of 0.99. As we are assuming a synchronized FL scenario, we simulate the parallel distributed learning by sequentially conducting local learning for the sampled clients and then aggregate them into a global model. The standard deviation is measured over 3 runs. The detailed learning setups for each datasets is provided in \autoref{tab:datasets}.

\begin{figure*}[ht!]
    \centering
    \includegraphics[width=0.99\textwidth]{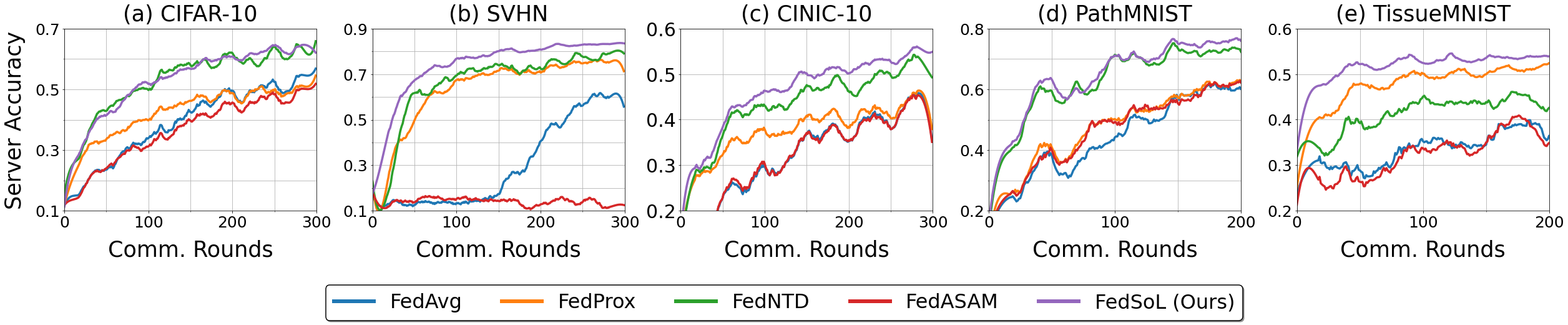}
    \caption{Learning curves of FL methods on LDA ($\alpha$=0.1). The curves are smoothed for clear visualization.}
    \vspace{-10pt}
    \label{fig:learning_curve}
\end{figure*}

\begingroup
\setlength{\tabcolsep}{2.0pt} 
\renewcommand{\arraystretch}{1.2}
\begin{table}[ht!]
\caption{Learning scenarios for each datasets.}
\centering
\begin{tabular}{cccc} 
\toprule
\multicolumn{1}{c}{\textbf{Datasets}} & \textbf{Clients} & \textbf{Comm. Rounds} & \textbf{Sampling Ratio}  \\ 
\hline\hline
MNIST             & 100              & 200                   & 0.1                      \\
CIFAR-10          & 100              & 300                   & 0.1                      \\
SVHN              & 100              & 200                   & 0.1                      \\
CINIC-10          & 200              & 300                   & 0.05                     \\
PathMNIST         & 200              & 200                   & 0.05                     \\
TissueMNIST       & 200              & 200                   & 0.05                     \\
\bottomrule
\end{tabular}
\label{tab:datasets}
\end{table}
\endgroup

\subsection{Algorithm Implementation Details}
We search for hyperparameters and select the best among the candidates. The hyperparameters for each method is provided in \autoref{tab:hyperparameters}. In the primary experiments, we use KL-divergence loss\,\citep{knowledge_distillation} with softened logits with temperature $\tau$=3 for the proximal loss for the adversarial weight perturbation in FedSOL.

\begingroup
\setlength{\tabcolsep}{1.5pt} 
\renewcommand{\arraystretch}{1.5}
\begin{table}[t!]
\caption{Algorithm-specific hyperparameters.}
\centering
\begin{tabular}{lll}
\toprule
\textbf{Methods} & \textbf{Selected} & \textbf{Searched Candidates} \\ 
\hline\hline
FedAvg  & None                                           & None \\
\hline
FedProx & $\mu$=1.0                                      &$\mu \in \{0.1, 0.5, 1.0, 2.0\}$\\
Scaffold & None                                          &None\\
FedNova  & None                                          &None\\
FedNTD   & $\beta$=1.0, $\tau$=1.0                           &$\beta \in \{0.5, 1.0\}$, $\tau \in \{1.0, 3.0\}$\\
FedSAM   & $\rho$=0.1                                    &$\rho \in \{0.1, 0.5, 1.0, 2.0\}$\\
FedASAM  & $\rho$=1.0                                    &$\rho \in \{0.1, 0.5, 1.0, 2.0\}$\\ 
FedDyn   &None                                           &None\\
MOON     & $\mu$=0.1, $\tau$=0.5                         &$\mu \in \{0.1, 0.5\}$, $\tau \in \{0.5, 1.0\}$\\
\hline
\textbf{FedSoL} & $\rho=2.0$  &$\rho \in \{0.1, 0.5, 1.0, 2.0\}$\\
\bottomrule
\end{tabular}
\label{tab:hyperparameters}
\end{table}
\endgroup

\section{Learning Curves}
\label{appendix_learning_curves}
To provide further insights into the learning process, we illustrate the learning curves of different FL methods in \autoref{fig:learning_curve}. Although we utilize different communication rounds for each dataset, the performance of the model becomes sufficiently saturated at the end of communication rounds. For all datasets, FedSoL not only achieves a superior final model at the end of the communication round but also demonstrates much faster convergence. Moreover, although some algorithms that perform well on a dataset fail on another\,(ex. FedNTD\,\citep{fedntd} underperforms compared to FedProx\,\citep{fedprox} on the TissueMNIST datasets), FedSOL consistently exhibits significant improvements when compared to the other baselines.

\section{Personalized Performance}
\label{appendix_pfl_performance}
In \autoref{tab:pFL}, we compare FedSOL with several methods specifically designed for personalized federated learning (pFL): PerFedAvg\,\citep{perfedavg}, FedBabu\,\citep{fedbabu}, and kNN-Per\,\citep{knnper}. Each method is assessed by fine-tuning them for $e$ local epochs from the global model after the final communication round. As global alignment is unnecessary for the personalized model, we fine-tune FedSOL using the local objective without perturbation and denote it as FedSOL-FT. The standard deviation is measured across the clients. The results reveal that our FedSOL-FT consistently outperforms other pFL methods under various scenarios. Furthermore, the gap is enlarged when local ($e$=1), implying that the global model obtained by FedSOL adapts more quickly to local distributions. We suggest that by integrating FedSOL with other methods specialized for pFL, we can attain superior performance for both the global server model and client local models.
\begingroup
\setlength{\tabcolsep}{3.7pt} 
\renewcommand{\arraystretch}{1.35}
\begin{table}[ht!]
\caption{Personalized FL performance after $\tau$ epochs of fine-tuning. The heterogeneity level is set as LDA ($\alpha = 0.1$).}
\centering
\small
\begin{tabular}{lcccc} 
\toprule
\multicolumn{1}{c}{\textbf{Method}} 
& $\boldsymbol{e}$ 
& \textbf{CIFAR-10}
& \textbf{SVHN}
& \textbf{TissueMNIST}\\ 
\hline\hline
Local-only          
& -             & 84.7${}_{\;\pm 12.8}$ & 87.4${}_{\;\pm 13.0}$ & 82.4${}_{\;\pm 15.5}$  \\
\hline
\multirow{2}{*}{FedAvg}
& 1             & 84.1${}_{\;\pm 13.4}$    & 86.6${}_{\;\pm 15.5}$ & 82.2${}_{\;\pm 17.5}$         \\
& 5             & 88.9${}_{\;\pm 8.9}\;\,$     & 92.1${}_{\;\pm 5.7}$\;\,  & 89.2${}_{\;\pm 10.1}$         \\ 
\hline
\multirow{2}{*}{PerFedAvg}           
& 1             & 80.5${}_{\;\pm 16.2}$  & 64.1${}_{\;\pm 30.3}$   & 82.3${}_{\;\pm 18.9}$         \\
& 5             & 86.3${}_{\;\pm 10.4}$  & 72.4${}_{\;\pm 21.2}$   & 88.8${}_{\;\pm 10.2}$         \\ 
\hline
\multirow{2}{*}{FedBabu}             
& 1             & 84.6${}_{\;\pm 12.7}$  & 88.7${}_{\;\pm 9.6}$    & 85.7${}_{\;\pm 14.3}$         \\
& 5             & 89.2${}_{\;\pm 8.4}$  & 92.7${}_{\;\pm 6.2}$     & 90.5${}_{\;\pm 8.8}$\;\,         \\ 
\hline
\multirow{2}{*}{kNN-Per}             
& 1             & 85.7${}_{\;\pm 12.3}$  & 86.4${}_{\;\pm 15.0}$   & 86.5${}_{\;\pm 14.2}$         \\
& 5             & 89.7${}_{\;\pm 8.1}$\;\,  & 92.8${}_{\;\pm 6.2}$\;\,     & 91.4${}_{\;\pm 7.5}$ \,\\ 
\hline
\multirow{2}{*}{\textbf{FedSoL\scriptsize{-FT} \scriptsize{(Ours)}}}         
& 1             & \underline{\textbf{87.5}}${}_{\;\pm 9.7}$\;\,  & \underline{\textbf{92.5}}${}_{\;\pm 7.4}$ \;\,      & \underline{\textbf{88.1}}${}_{\;\pm 12.2}$       \\
& 5             & \underline{\textbf{90.5}}${}_{\;\pm 7.8}$\;\,  & \underline{\textbf{95.0}}${}_{\;\pm 3.9}$\;\,       & \underline{\textbf{91.6}}${}_{\;\pm 6.9}$\;\,       \\
\bottomrule
\end{tabular}
\label{tab:pFL}
\end{table}
\endgroup

\section{Performance on Larger Datasets}
\label{appendix_cifar100_imagenet100}
In \autoref{tab:main_std}, we show that FedSOL consistently achieves performance gains, while most existing FL (Federated Learning) methods are sensitive to learning setups. Although FedNTD marginally outperforms FedSOL in some CIFAR-10 cases, it significantly falls behind in most others. In \autoref{tab:cifar100_imagenet100}, we conducted experiments on the CIFAR-100 and ImageNet-100 datasets. We used the ResNet-18 model, distributing each dataset across 100 clients with a sampling ratio of 0.1 and optimized for 5 local epochs. Our observations indicate that FedSOL maintains its effectiveness in both datasets, whereas FedNTD's performance decreases in CIFAR-100.

\begingroup
\setlength{\tabcolsep}{6.0pt} 
\renewcommand{\arraystretch}{1.1}
\begin{table}[ht!]
\small
\caption{Test Accurac on CIFAR-100 and ImageNet-100.}
\label{tab:cifar100_imagenet100}
\begin{tabular}{lcccc}
\toprule
\multicolumn{1}{c}{\multirow{2}{*}{\textbf{Method}}} & \multicolumn{3}{c}{\textbf{CIFAR-100}}                                           & \multicolumn{1}{c}{\textbf{ImageNet-100}} \\
                                 & \multicolumn{1}{c}{s\,=\,5} & \multicolumn{1}{c}{s\,=\,10} & \multicolumn{1}{c}{$\alpha$\,=\,0.05} & \multicolumn{1}{c}{$\alpha$\,=\,0.1}                 \\
\hline\hline
FedAvg                           
& 42.43\,\,\,                   
& 53.23\,\,\,                      
& 48.69\,\,\,                     
& 43.41\,\,\,                                       \\
FedProx                          
& 39.03\,\footnotesize{\textcolor{red}{$\downarrow$}}                   
& 48.38\,\footnotesize{\textcolor{red}{$\downarrow$}}                      
& 46.75\,\footnotesize{\textcolor{red}{$\downarrow$}}                     
& 34.49\,\footnotesize{\textcolor{red}{$\downarrow$}}                                     \\
FedNTD                           
& 39.32\,\footnotesize{\textcolor{red}{$\downarrow$}}                   
& 52.23\,\footnotesize{\textcolor{red}{$\downarrow$}}                      
& 48.35\,\footnotesize{\textcolor{red}{$\downarrow$}}                     
& 44.08\,\footnotesize{\textcolor{green!50!black}{$\uparrow$}}                                   \\
\hline
\textbf{FedSOL}                    
& \textbf{44.21}\,\footnotesize{\textcolor{green!50!black}{$\uparrow$}}                   
& \textbf{53.78}\,\footnotesize{\textcolor{green!50!black}{$\uparrow$}}                      
& \textbf{49.25}\,\footnotesize{\textcolor{green!50!black}{$\uparrow$}}                     
& \textbf{44.97}\,\footnotesize{\textcolor{green!50!black}{$\uparrow$}}
\\
\bottomrule
\end{tabular}
\vspace{-12pt}
\end{table}
\endgroup

\section{Head Perturbation in Larger Models} 
\label{appendix_partial_perturbation_larger}

To validate the effectiveness of partial perturbation strategy, we extended the comparison experiment in \autoref{tab:partial_perturb} to \autoref{tab:partial_perturbation}. The results on VggNet-11 and ResNet-18 indicate that perturbing only the last classifier layer (\textit{head}) is almost as effective as perturbing the entire model (\textit{full}), saving significant computational cost.
\begingroup
\setlength{\tabcolsep}{4.0pt} 
\renewcommand{\arraystretch}{1.1}
\small
\begin{table}[ht!]
\caption{Effect of partial perturbation on CIFAR-10 ($\alpha$=0.1).}
\label{tab:partial_perturbation}
\begin{tabular}{lccc}
\toprule
\multicolumn{1}{c}{\textbf{Model}} & \multicolumn{1}{c}{\textbf{FedAvg}} & \multicolumn{1}{c}{\textbf{FedSOL (\textit{full})}} & \textbf{FedSOL (\textit{head})} \\
\hline\hline
VggNet-11
& 41.30 
& 56.44 \textbf{\small{\textcolor{green!60!black}{(+\,15.14)}}}  
& 56.39 \textbf{\small{\textcolor{green!60!black}{(+\,15.09)}}} \\
ResNet-18      
& 49.92 
& 66.69 \textbf{\small{\textcolor{green!60!black}{(+\,16.77)}}} 
& 66.32 \textbf{\small{\textcolor{green!60!black}{(+\,16.04)}}}
\\
\bottomrule
\end{tabular}
\end{table}
\endgroup

\section{Comparison to the Sharpness-Aware Optimization}
\label{appendix_discussion_on_sam}
\begin{figure*}[ht!]
    \vspace{-4pt}
    \centering
    \includegraphics[width=0.98\textwidth]{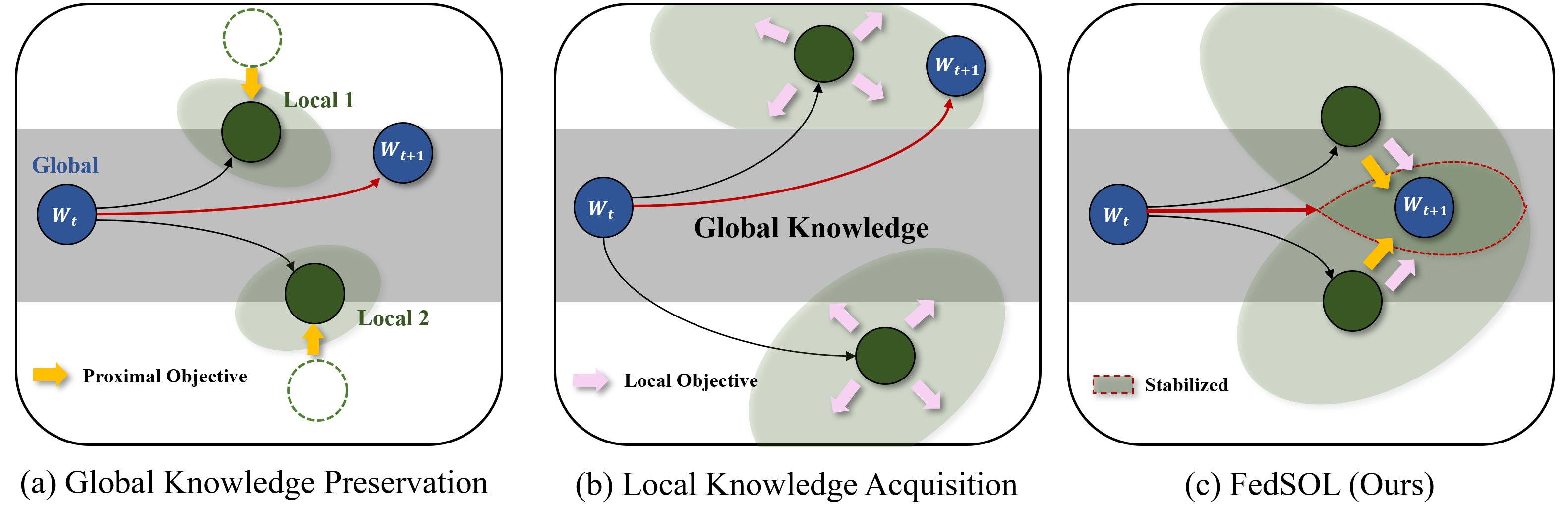}
\caption{An overview of FL scenarios. \textbf{\textcolor{white!40!black}{Gray-shaded}} areas represent global knowledge, while \textbf{\textcolor{green!40!black}{Green-shaded}} areas represent local knowledge. (a) Learning with the proximal objective achieves global knowledge preservation but limits local knowledge acquisition. (b) Learning with the local objective effectively acquires local knowledge but results in forgetting global knowledge. In (c), the orthogonal learning strategy of FedSOL stabilizes the learning process by resolving conflicts between these two objectives.}
\label{fig:overview}
\vspace{-2pt}
\end{figure*}

\subsection{SAM Optimization in FL} Recent studies have begun to suggest that enhancing local learning generality can significantly boost FL performance\,\citep{fedalign, fedsam, fedasam}, aiding the global model in generalizing more effectively. Inspired by the latest findings that connect loss geometry to the generalization gap\,\citep{swa, large_batch_generalization_gap, fantastic, entropy_sgd}, those works seek for \textit{flat minima}, utilizing the recently proposed Sharpness-Aware Minimization (SAM)\,\citep{sam} as the local optimizer. For instance, FedSAM\,\citep{fedsam} and FedASAM\,\citep{fedasam} demonstrate the benefits of using SAM and its variants as local optimizer. Meanwhile, FedSMOO\,\citep{fedsmoo} incorporates a global-level SAM optimizer, and FedSpeed,\citep{fedspeed} employs multiple gradient calculations to encourage global consistency. By improving local generality, these approaches mitigate the conflicts between individual local objectives, contributing to the overall smoothness of the aggregated global model\,\citep{fedsmoo, fedsam, fedasam}.

\subsection{Limitations of using SAM in FL}
Although these approaches have demonstrated competitive performance without proximal restrictions, their ability to generalize effectively within their respective local distributions does not necessarily guarantee the preservation of previous global knowledge during local learning. This is due to the inherent conflict between global and local objectives. In our FedSOL, we introduce the use of proximal perturbation as a means to achieve an orthogonal local gradient, which does not contribute to an increase in proximal loss. This approach can be understood as implicitly incorporating the effect of proximal restriction into SAM, achieved by adjusting the perturbation's direction and magnitude during local learning. It is also worth mentioning that the effect of proximal perturbation depends on the relationship between the local and proximal objectives. In the extreme case where the local objective is identical to the proximal objective, our FedSOL collapses into the original SAM.

In \autoref{fig:overview}, we illustrate a conceptual overview of the global and local knowledge trade-off in FL. In \autoref{fig:overview}\textcolor{red}{(a)} and \autoref{fig:overview}\textcolor{red}{(b)}, learning on one objective undermines the effect of the other. However, the orthogonal learning of FedSOL stabilizes the local learning by tackling the conflicts between the two objectives\,(\autoref{fig:overview}\textcolor{red}{(c)}).

\begin{figure*}[t!]
    \centering
    \includegraphics[width=0.9\textwidth]{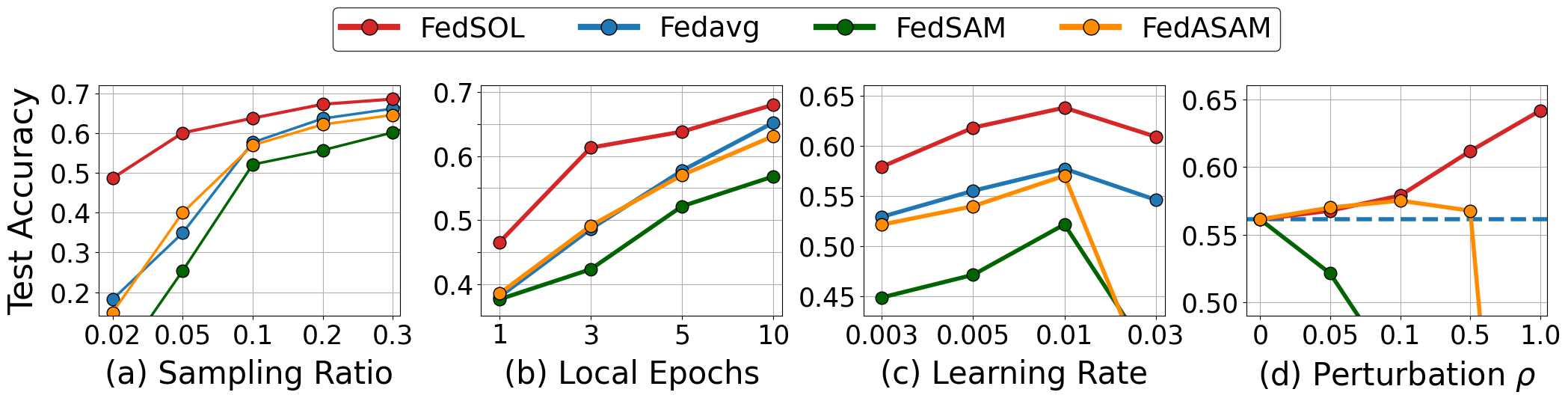}
\caption{{Performance on CIFAR-10 ($\alpha$=0.1) datasets.}}
\label{fig:ablation_R}
\vspace{-4pt}
\end{figure*}

\section{Other Perturbation Strategies}
\label{appendix_perturb_strategy}

In our work, we propose the use of proximal perturbation as our primary strategy. This section compares various perturbation strategies and discusses the effectiveness of using the proximal gradient for weight perturbation. A straightforward approach would involve using a linear combination of the local objective, which includes both the local loss  $\mathcal{L}^{k}_{\mathrm{local}}$ and the proximal loss $\mathcal{L}^{k}_{\mathrm{p}}$ in \autoref{eq_proximal_restriction}. This combination is applied in SAM-like optimization\,(\autoref{eq_sam_minmax}) as follows:
\begin{equation}
    \underset{\bwk}{\text{min}} \; \underset{{\| \boldsymbol{\epsilon} \|}_{2} < \rho }{\text{max}}\; \big[\mathcal{L}_{\mathrm{local}}^k(\bwk) + \beta\cdot\mathcal{L}_p^k(\bwk;\bwg)\big]\,.
\end{equation}
In the above equation, the gradients for weight perturbation and parameter update are obtained from the same objective. 

However, this approach encounters the same drawbacks as when using each method on its own. The combined loss also varies considerably across clients due to heterogeneous local distributions, causing the smoothness to largely rely on individual local distributions. Furthermore, the negative correlation between the gradients of the two objectives within the combined loss still limits local learning. Consequently, this approach neither preserves global knowledge from proximal objective nor effectively acquires the local knowledge as desired. 

Instead in FedSOL, we overcome this issue by decoupling this directly combined loss into the proximal loss $\mathcal{L}^{k}_{\mathrm{p}}$ for weight perturbation and the local loss $\mathcal{L}^{k}_{\mathrm{local}}$ for weight updates. To further analyze the relationship between loss functions and weight perturbation in FedSOL optimization, we conduct an ablation study on the following strategies.

\vspace{2pt}
\begin{itemize}
    \item $A_0$: Use local loss without any weight perturbation \textbf{(FedAvg)}.
    \vspace{3pt}
    \item $A_1$: Use local loss, but get the local loss gradient at weights perturbed by the proximal gradient \textbf{(FedSOL)}.
    \item $A_2$: Use combined loss, but get the proximal loss gradient at weights perturbed by the proximal gradient.
    \vspace{3pt}
    \item $A_3$: Use combined loss, but get the proximal loss gradient at weights perturbed by the proximal gradient.
    \vspace{3pt}
    \item $A_4$: Use combined loss, but get the combined loss gradient at weights perturbed by the combined gradient.
    \vspace{3pt}
    \item $A_5$: Use combined loss without any weight perturbation \textbf{(Proximal Restriction)}.
    \vspace{3pt}
    \item $A_6$: Use combined loss, but get the local loss gradient at proximally perturbed weight loss.
    \vspace{2pt}
\end{itemize}

\vspace{6pt}
\noindent We exclude the strategies that obtaining proximal loss at the perturbed weights using the local loss gradient i.e., $\mathcal{L}_{p}(\bwk+\boldsymbol{\epsilon}_{c}^{*}$), where $\boldsymbol{\epsilon}_{c}^{*}$ $=$ $\rho\,\frac{\boldsymbol{g}_p+\boldsymbol{g}_l}{\lVert \boldsymbol{g}_p + \boldsymbol{g}_l \rVert}$, as it leads the learning to diverge. The detailed formulation for each method is provided in Table \ref{tab:ablation_objectives} with its corresponding performance. The results in \autoref{tab:ablation_objectives} demonstrates that utilizing the local loss gradient at weights perturbed by the proximal loss gradient ($A_1$ in \autoref{tab:ablation_objectives}) yields outperforms the other approaches. We suggest that our FedSOL is an effective way to integrate proximal restriction effect into SAM optimization in FL.

\begingroup
\setlength{\tabcolsep}{1.0pt} 
\renewcommand{\arraystretch}{1.3}
\begin{table}[ht!]
\caption{Detailed formulation for each method and their performance on CIFAR-10 datasets (LDA $\alpha$=0.1).}
\centering
\begin{tabular}{clc} 
\toprule
\textbf{Name} &\multicolumn{1}{c}{\textbf{Method Formulation}}&\multicolumn{1}{c}{\textbf{Performance}}    \\ 
\hline\hline
$A_0$            
& $\mathcal{L}_{\text{local}}(\bwk)$ & 56.13 \\
$A_1$            
& $\mathcal{L}_{\text{local}}(\bwk + \boldsymbol{\epsilon}_p^{*})$ & \textbf{64.13} \\
$A_2$            
& $\mathcal{L}_{\text{local}}(\bwk) + \beta\cdot\mathcal{L}_{p}(\bwk + \boldsymbol{\epsilon}_p^{*})$ & 53.85\\
$A_3$            
& $\mathcal{L}_{\text{local}}(\bwk + \boldsymbol{\epsilon}_p^{*}) + \beta\cdot\mathcal{L}_{p}(\bwk + \boldsymbol{\epsilon}_p^{*})$ & 60.28\\
$A_4$            
& $\mathcal{L}_{\text{local}}(\bwk +  \boldsymbol{\epsilon}_{\text{c}}^{*}) + \beta\cdot\mathcal{L}_{p}(\bwk +  \boldsymbol{\epsilon}_{\text{c}}^{*})$ & 45.72\\
$A_5$            
& $\mathcal{L}_{\text{local}}(\bwk) + \mathcal{L}_{p}(\bwk)$  & 61.76\\
$A_6$            
& $\mathcal{L}_{\text{local}}(\bwk + \boldsymbol{\epsilon}_p^{*}) + \beta\cdot\mathcal{L}_{p}(\bwk)$ & 44.12\\
\bottomrule
\end{tabular}
\label{tab:ablation_objectives}
\end{table}
\endgroup


\section{Client-side Computational Cost}
\label{appendix_computational_cost}
In \autoref{tab:partial_perturb}, we analyze the FLOPS required for FedSOL and note that perturbing only the head part is almost as effective as full perturbation, yet it requires only 33\% more computation than FedAvg. Although FedSOL necessitates backward computation twice, this does not lead to increased GPU memory usage, as FedSOL does not store these gradients simultaneously. The slight increase in memory usage ($< 0.01$\%) arises from calculating the adaptive perturbation strength for each layer. In \autoref{tab:gpu_usage}, we present the latency for a single local step with the ImageNet-100 dataset, using a ResNet-18 model on an NVIDIA A6000 GPU. We measured FedSOL's latency using an L2 proximal loss for a fair comparison. While FedSOL involves additional local computation, we emphasize that in FL, the energy consumed for communication typically surpasses that for computation. The faster convergence of FedSOL substantially reduces the total energy consumption. In the ImageNet-100 experiment detailed in \autoref{tab:cifar100_imagenet100}, for example, FedSOL achieves FedAvg's 300-round performance by round 238.

\begingroup
\setlength{\tabcolsep}{3.0pt} 
\renewcommand{\arraystretch}{1.1}
\begin{table}[ht!]
\small
\caption{Measured latency for a single local step.}
\vspace{-7.5pt}
\label{tab:gpu_usage}
\centering
\begin{tabular}{lccccc} 
\toprule
\multicolumn{1}{c}{\textbf{Usage}}     
& \textbf{FedAvg} & \textbf{FedProx} & \textbf{FedNTD} & \textbf{FedSAM} & \textbf{FedSOL}  \\ 
\hline\hline
Latency 
& 2.846\,s    
& 2.865\,s     
& 2.996\,s   
& 3.380\,s    
& 2.900\,s     \\
\bottomrule
\end{tabular}
\end{table}
\endgroup

\section{Effect of Learning Factors on FedSAM}
\label{appendix_learning_factors_fedsam_fedasam}

\autoref{fig:ablation_R} presents the impact of learning factors on FedSAM\,\citep{fedsam} and FedASAM\,\citep{fedasam}, showing that both methods are more sensitive to these factors compared to FedSOL. In most of our experiments, SAM-related FL methods shows inferior performance compared to FedAvg. This may be because SAM, aiming to enhance local model generality, becomes less effective for small models, under conditions of high data heterogeneity, or with a low sampling ratio. Future research is expected to identify the conditions where employing SAM on the local side becomes beneficial.

\clearpage
\onecolumn
\newcommand{\bv}{\boldsymbol{v}}
\section{Proof of Proposition}
\label{appendix_proof}
The notion of $\approx$ in the main papers for (Especially for \autoref{eq_sol_taylor} and \autoref{eq_loss_difference_definition}) are supported by Taylor's theorem. We will use the following formulation for $C^2$ functions. All matrix norms are the largest singular value.

\begin{theorem}[Taylor's theorem]
\label{taylorthm}
If $f$ is $C^2$ function at the open ball contains $\bw$ and $\bw+\bv$, we have:
\[
f(\bw+\bv) = f(\bw) +  \langle \nabla f(\bw) ,\; \bv \rangle + R_f(\bv;\bw)\:,\:\text{where}\:\:|R_f(\bv\,;\bw)| \leq \frac{1}{2}\|\bv\|^2 \max_{t\in[0,1]} \| \nabla^2 f (\bw+ t \bv) \|_2\,.
\]
For $R (\bv\,;\bw)$, there are two well-known representations:
\begin{itemize}
\item There exists $t \in (0,1)$ such that $\displaystyle R_f(\bv\,;\bw) = \frac{1}{2} \bv^\top \nabla^2 f(\bw + t\bv) \bv $,
\item $\displaystyle R_f(\bv\,;\bw) = \int_0^1 (1-t) \bv^\top \nabla^2 f(\bw + t\bv)\, \bv\, dt $.
\end{itemize}
\end{theorem}
\vspace{2pt}
Now, We first precise the notion of $\approx$ in the \autoref{eq_sol_taylor}:   
\begin{align}
\label{eq_9_precise}
\Delta^{\mathrm{FedSOL}}\mathcal{L}^k(\bwk) &= \mathcal{L}^k\big(\bwk - \gamma\,\boldsymbol{g}_u^{\mathrm{FedSOL}}(\bwk)\big) - \mathcal{L}^k(\bwk)\notag\\
&= -\gamma \langle \nabla_{\bwk}  \mathcal{L}^k(\bwk),\; \boldsymbol{g}_u^{\mathrm{FedSOL}}(\bwk)\rangle + R_{\mathcal{L}^k} (-\gamma\,\boldsymbol{g}_u^{\mathrm{FedSOL}}(\bwk)\,;\bwk)\,,
\end{align}
and \autoref{eq_loss_difference_definition}:
\begin{align}
\label{eq_10_precise}
\boldsymbol{g}_u^{\mathrm{FedSOL}}(\bwk)&={\nabla}_{\bwk} \mathcal{L}^k_{\mathrm{local}}(\bwk+\boldsymbol{\epsilon}^{*}_p) \notag\\
&= \boldsymbol{g}_l(\bwk) + \rho\, \nabla^2_{\bwk} \mathcal{L}^k_{\mathrm{local}}(\bwk)\hat{\boldsymbol{g}}_p(\bwk)+ \boldsymbol{R}_{{\nabla}_{\bwk} \mathcal{L}^k_{\mathrm{local}}}(\rho \hat{\boldsymbol{g}}_p(\bwk)\,;\bwk) \,,
\end{align}
where $\boldsymbol{R}_{{\nabla} \mathcal{L}^k_{\mathrm{local}}}(\rho \hat{\boldsymbol{g}}_p(\bwk)\,;\bwk)$ is a vector, where the $i$-th value is ${R}_{\partial_i \mathcal{L}^k_{\mathrm{local}}}(\rho \hat{\boldsymbol{g}}_p(\bwk)\,;\bwk)$, which is the residual term with $i$-th directional derivative $\partial_i \mathcal{L}^k_{\mathrm{local}}$.\\
\vspace{5pt}

\noindent By substituting the above expression for $\boldsymbol{g}_u^{\mathrm{FedSOL}}(\bwk)$ into $\Delta^{\mathrm{FedSOL}}\mathcal{L}^k(\bwk)$, we have:
\begin{align*}
&\Delta^{\mathrm{FedSOL}}\mathcal{L}^k_{\{\mathrm{local},p\}}(\bwk) = -\gamma \langle \nabla_{\bwk}  \mathcal{L}^k_{\{\mathrm{local},p\}}(\bwk), \boldsymbol{g}_u(\bwk)\rangle + R_{\mathcal{L}^k_{\{\mathrm{local},p\}}} (-\gamma\,\boldsymbol{g}_u^{\mathrm{FedSOL}}(\bwk)\,;\bwk)\,,\notag \\
&= -\gamma \left( \langle \nabla_{\bwk}  \mathcal{L}^k_{\{\mathrm{local},p\}}(\bwk),\boldsymbol{g}_l \rangle + \rho\, \langle \nabla_{\bwk}  \mathcal{L}^k_{\{\mathrm{local},p\}}(\bwk),\nabla^2 \mathcal{L}^k_{\mathrm{local}}\, \hat{\boldsymbol{g}}_p \rangle \right) +\\
&\quad\:\: \underbrace{-\gamma \langle \nabla_{\bwk}  \mathcal{L}^k_{\{\mathrm{local},p\}}(\bwk), \boldsymbol{R}_{{\nabla} \mathcal{L}^k_{\mathrm{local}}}(\rho \hat{\boldsymbol{g}}_p(\bwk)\,;\bwk)
\rangle + R_{\mathcal{L}^k_{\{\mathrm{local},p\}}} (-\gamma\,\boldsymbol{g}_u^{\mathrm{FedSOL}}(\bwk)\,;\bwk)}_{\mathcal{E}^k_{\{\mathrm{local},p\}}}\,,
\end{align*}
where $\mathcal{E}^k_{\{\mathrm{local},p\}}$ is the total residual term:
\[
\mathcal{E}^k_{\{\mathrm{local},p\}} = -\gamma  \sum_i \partial_i \mathcal{L}^k_{\{\mathrm{local},p\}}(\bwk) {R}_{\partial_i \mathcal{L}^k_{\mathrm{local}}}(\rho \hat{\boldsymbol{g}}_p(\bwk)\,;\bwk) + R_{\mathcal{L}^k_{\{\mathrm{local},p\}}} (-\gamma\,\boldsymbol{g}_u^{\mathrm{FedSOL}}(\bwk)\,;\bwk)\,.
\]

\noindent For representing the magnitude of residual term effectively, we will assume three constants. It is important to note that the constants can be made smaller by concentrating on the optimization-relevant region rather than the entire weight space.

\begin{assumption}
Matrix norm of Hessian and the norm of gradient for $\mathcal{L}^k_{\mathrm{local}}$ are bounded:
\[
D^k = \sup_{\bw} \| \nabla^2 \mathcal{L}^k_{\mathrm{local}}(\bw) \| < \infty\,,\,B^k = \sup_{\bwk} \|  \nabla_{\bwk} \mathcal{L}^k_{\mathrm{local}}(\bwk) \| < \infty\,.
\]
\end{assumption}

\begin{assumption}
For any linear path connecting $\bw$ and $\bv$ with length $\rho$, we define the following coefficient:
\[
C^k_{\rho, \{\mathrm{local},p\}} = \sup_{\bw, \bv, \|\bw-\bv\|=\rho}\,\max_{t\in[0,1]}\left\| \sum_i  \partial_i \mathcal{L}^k_{\{\mathrm{local},p\}}(\bw)\nabla^2 \partial_i \mathcal{L}^k_{\mathrm{local}}(\bw + t (\bv-\bw ))  \right\| < \infty\,.
\]
\end{assumption}

\noindent The first term becomes:
\[
 -\gamma \rho^2 \int_0^1 (1-t) \hat{\boldsymbol{g}}_p(\bwk)^\top  \left( \sum_i  \partial_i \mathcal{L}^k_{\{\mathrm{local},p\}}(\bwk)\nabla^2 \partial_i \mathcal{L}^k_{\mathrm{local}}(\bw + t \rho \hat{\boldsymbol{g}}_p(\bwk) )\right) \, \hat{\boldsymbol{g}}_p(\bwk)\, dt \,,
\]
and we can easily see that magnitude of this term can be bounded with $\frac{1}{2}\gamma \rho^2 C^k_{\rho, \{ \mathrm{local},p\}}$, by \autoref{taylorthm}. By same procedure, it is easy to see the second term is bounded by $ \frac{1}{2} \gamma^2 D^k (B^k)^2$. Consequently, we can conclude:
\begin{equation}
\label{pfeq1}
\Delta^{\mathrm{FedSOL}}\mathcal{L}^k_{\{\mathrm{local},p\}}(\bwk) =\gamma \left( \big\langle \nabla_{\bwk}  \mathcal{L}^k_{\{\mathrm{local},p\}}(\bwk),\boldsymbol{g}_l \big\rangle + \rho\, \big\langle \nabla_{\bwk}  \mathcal{L}^k_{\{\mathrm{local},p\}}(\bwk),\nabla^2 \mathcal{L}^k_{\mathrm{local}}\, \hat{\boldsymbol{g}}_p \big\rangle \right) + \mathcal{E}_{\{\mathrm{local},p\}} \,,
\end{equation}
where $\mathcal{E}_{\{\mathrm{local},p\}}$ has magnitude:
\[
\mathcal{E}^k_{\{\mathrm{local},p\}} = \mathcal{O}\Big(\gamma \rho^2 C^k_{\rho, \{ \mathrm{local},p\}} +\gamma^2 D^k (B^k)^2\Big)\,.
\]

\noindent \textbf{Remark}. In this analysis section, to align with the adaptive perturbation scheme utilized in real experiments, we cautiously suggest considering the constant strength as $\rho = 2.0 / \sqrt{\text{\# params}}$ in terms of order. In the FedSOL algorithm, adaptive perturbation strength is employed, as detailed in Section~\ref{subsec:adaptive_perturbation}. Here, from  \autoref{eq:adaptive_perturbation}, the squared sum of perturbation strength is $\rho^2$, substantially lower than in scenarios assuming a constant strength $\rho$ (where the squared sum would be $\rho^2 \times (\text{\# params})$). For simplicity, our analysis primarily focuses on scenarios with constant perturbation strength. Consequently, within this analysis, the effective perturbation strength should be considerably lower than the experimental setting of $\rho=2.0$ in FedSOL. That is, it should be treated as the order of $2.0/\sqrt{\text{\# params}}$ to match the parameter-wise squared sum of perturbation strengths.

\subsection{Proof of Proposition 1} 
Regarding Proposition 1, from \autoref{pfeq1}, we obtain:
\begin{align*}
\Delta^{\mathrm{FedSOL}} \mathcal{L}^k_{p} &= -\gamma \left( \langle \boldsymbol{g}_p,\boldsymbol{g}_l \rangle + \rho\, \langle  \boldsymbol{g}_p,\nabla^2 \mathcal{L}^k_{\mathrm{local}}\, \hat{\boldsymbol{g}}_p \rangle \right) + \mathcal{E}^k_{p} \\
&=  -\gamma\Big(\langle \boldsymbol{g}_l\,,\boldsymbol{g}_p \rangle + \rho \cdot \hat{\boldsymbol{g}}_p{\!\!}^\top \nabla^2\mathcal{L}^k_{\mathrm{local}}\,{\boldsymbol{g}_p}\Big) + \mathcal{O}\Big(\gamma \rho^2 C^k_{\rho, p} +\gamma^2 D^k (B^k)^2\Big) \,.
\end{align*}
Furthermore, if the $\nabla^2\mathcal{L}^k_{\mathrm{local}}(\bwk)$ is positive semi-definite, $\boldsymbol{g}_p{\!\!}^\top$ $ \nabla^2\mathcal{L}^k_{\mathrm{local}}$ $\,{\boldsymbol{g}_p} \geq 0$, and we can guarantee that the second term is nonnegative as well.

\subsection{Proof of Proposition 2}
Similarly, from \autoref{pfeq1}, we derive:
\begin{align*}
\Delta^{\mathrm{FedSOL}}\mathcal{L}^k_{\mathrm{local}}(\bwk) &= -\gamma \left( \langle \nabla_{\bwk}  \mathcal{L}^k_{\mathrm{local}}(\bwk),  \nabla_{\bwk}  \mathcal{L}^k_{\mathrm{local}}(\bwk) \rangle + \rho\, \langle \nabla_{\bwk}  \mathcal{L}^k_{\mathrm{local}}(\bwk),\nabla_{\bwk} ^2 \mathcal{L}^k_{\mathrm{local}}\, \hat{\boldsymbol{g}}_p \rangle \right) + \mathcal{E}^k_{\mathrm{local}} \\
& =  -\gamma \left( \|\nabla_{\bwk}  \mathcal{L}^k_{\mathrm{local}}(\bwk) \|^2  + \frac{\rho}{2}\, \langle \hat{\boldsymbol{g}}_p, \underline{2\nabla_{\bwk}^2 \mathcal{L}^k_{\mathrm{local}} \nabla_{\bwk}  \mathcal{L}^k_{\mathrm{local}}(\bwk)} \rangle\right) + \mathcal{O}\Big(\gamma \rho^2 C^k_{\rho, \mathrm{local}} +\gamma^2 D^k (B^k)^2\Big)\\
& = -\gamma \left( \|\nabla_{\bwk}  \mathcal{L}^k_{\mathrm{local}}(\bwk) \|^2  + \underline{\nabla_{\bwk} \|\nabla_{\bwk}  \mathcal{L}^k_{\mathrm{local}}(\bwk) \|^2}  \cdot \frac{\rho}{2}\, \hat{\boldsymbol{g}}_p \right) + \mathcal{O}\Big(\gamma \rho^2 C^k_{\rho, \mathrm{local}} +\gamma^2 D^k (B^k)^2\Big)
\end{align*}

\noindent For the FedAvg update of $\mathcal{L}^k_{\mathrm{local}}(\bwk)$, as derived in \autoref{eq_9_precise}, we have:
\begin{align}
\label{pfeq2}
\Delta^{\mathrm{FedAvg}}\mathcal{L}^k_{\mathrm{local}}(\bwk) &= -\gamma \langle \nabla_{\bwk}  \mathcal{L}^k(\bwk), \boldsymbol{g}_l \rangle + R_{\mathcal{L}^k_{\mathrm{local}}}(-\gamma\,\boldsymbol{g}_l(\bwk)\,;\bwk)\notag\\
&= -\gamma \|\nabla  \mathcal{L}^k_{\mathrm{local}}(\bwk) \|^2  + \mathcal{O}\big(\gamma^2 D^k (B^k)^2\big) \,.
\end{align}
On the other hand, from \autoref{pfeq2}, applying the first-order Taylor approximation to $\bwk \mapsto \|\nabla \mathcal{L}^k_{\mathrm{local}}(\bwk)\|^2 $, we have:
\begin{align*}
&\Delta^{\mathrm{FedAvg}}\mathcal{L}^k_{\mathrm{local}}\left(\bwk + \frac{\rho}{2} \hat{\boldsymbol{g}}_p \right) = -\gamma \left\|\nabla_{\bwk}   \mathcal{L}^k_{\mathrm{local}}\left(\bwk + \frac{\rho}{2} \hat{\boldsymbol{g}}_p \right)\right\|^2  + \mathcal{O}\big(\gamma^2 D^k (B^k)^2\big)\\
& = \gamma \left( \|\nabla_{\bwk}  \mathcal{L}^k_{\mathrm{local}}(\bwk) \|^2  +  \nabla_{\bwk} \left\|\nabla_{\bwk}  \mathcal{L}^k_{\mathrm{local}}(\bwk) \right\|^2  \cdot \frac{\rho}{2}\, \hat{\boldsymbol{g}}_p  \right) + \mathcal{O}\big( \gamma\rho^2  
C^k_{0, \mathrm{local}} + \gamma\rho^2 (D^k)^2 + \gamma^2 D^k (B^k)^2\big) \,.
\end{align*}
Therefore, $\Delta^{\mathrm{FedAvg}}\mathcal{L}^k_{\mathrm{local}}$ $(\bwk + \frac{\rho}{2} \hat{\boldsymbol{g}}_p)$ is equivalent with $\Delta^{\mathrm{FedSOL}}\mathcal{L}^k_{\mathrm{local}}(\bwk)$, up to order:
\[
\mathcal{O}\Big( \gamma \rho^2 C^k_{\rho, \mathrm{local}}+ \gamma\rho^2  
C^k_{0, \mathrm{local}} + \gamma\rho^2 (D^k)^2 + \gamma^2 D^k (B^k)^2 \Big)\,.
\]

\section{Toy Example for Explaining FedSOL}

Consider a two-dimensional weight space, $\mathbb{R}^2$, and denote the weights as $(u,v)\in\mathbb{R}^2$. We define the local loss function $\mathcal{L}_{\mathrm{local}}$ and the proximal loss function $\mathcal{L}_p$. The local minimum is represented as $(u_l, v_l)$, and the aggregated server weight is denoted as $(0,0)$. The most intuitive form of losses that can be considered is:
\[
\mathcal{L}_{\mathrm{local}}(u,v) = \frac{1}{2} (u-u_l)^2 + \frac{\delta}{2}(v-v_l)^2 \:\mathrm{and}\: \mathcal{L}_p(u,v) = \frac{\mu}{2} ( u^2 + v^2)\,,
\]
where $\delta$ is a positive coefficient. A value of $\delta$ indicates that the local objective is more influenced by the variable $u$ than by the direction $v$ if $\delta<1$ and vice versa. $\delta$ can be also interpreted as the proportion of one class over another in binary classification.
\vspace{2pt}

\noindent Since $\mathbf{\epsilon}^*_p = \frac{(u,v)}{\sqrt{u^2 + v^2}}$, the FedSOL gradient on local side becomes:
\[
g^{\mathrm{FedSOL}} = \left( (u-u_l) +  \rho \frac{u}{\sqrt{u^2 + v^2}}\,, \delta \cdot \Big( (v-v_l) +  \rho \frac{v}{\sqrt{u^2 + v^2}}  \Big)  \right)\,,
\]
while with the adding proximal loss term (like FedProx), we have an overall local gradient:
\[
g^{\mathrm{FedProx}} = \left( (u-u_l) +  \mu \cdot u , \delta \cdot (v-v_l) +   \mu \cdot v \right)\,.
\]
\noindent When considering proximal loss, proximal regularization is applied in the $u$ and $v$ directions without considering local loss, which can be sub-optimal. The sub-optimality can be summarized in the sense of global alignment and local learnablity and this can be resolved with FedSOL, which uses local gradient update in the proximal-loss-sensitive point, while not harming local learnability as \autoref{prop2:local_change}. This argument strengthen the discussion for effectiveness of reducing negative inner product in \autoref{prop1:proximal_change}.

\noindent The equilibrium point in both algorithms is defined where the gradient equals zero, and in the current setting, it is unique. We investigate each algorithm on this setting as follows:

\subsection*{Proximal Regularization (FedProx)}

\textbf{Analysis}: In FedProx, the learned weight $(u^*, v^*)$ is calculated as $\left( \frac{1}{1+\mu} u_l,  \frac{\delta}{\delta + \mu} v_l \right)$. This reflects the influence of both the strength of normalization and the local curvature. Furthermore, the resulting vector from $(0,0)$ to $(u^*, v^*)$ is biased towards the direction with greater curvature compared to $(u_l, v_l)$.

\noindent \textbf{Implication}: While this directional bias may not pose significant issues in local learning, it becomes problematic in FL where weights from different clients are averaged. Since each client may have a different $\delta$, the result of normalization influenced by local client loss might not be optimal in an FL context.

\subsection*{FedSOL}
\textbf{Analysis}: FedSOL introduces a normalization that maintains the same direction but reduces the vector magnitude as $\rho$. 
\begin{proof}
Let us consider the polar coordinate of $(u^*,v^*)$ and $(u_l, v_l)$ as $(r^*,\theta^*)$ and $(r_l, \theta_l)$ respectively. Furthermore, $(u^*, v^*)$ satisfies:
\[
(u^* - u_l, v^*-v_l) = \left( -\rho \frac{u^*}{\sqrt{(u^*)^2 + (v^*)^2}}, -\rho \frac{v^*}{\sqrt{(u^*)^2 + (v^*)^2}} \right)\,.
\]
Then $\theta^*=\theta_l=\theta$ is ensured by $\frac{v_l}{u_l} = \frac{v^*}{u^*}$. Now, the above equation gives $r^* = r_l - \rho$.
\end{proof}

\noindent \textbf{Benefit}: This implies that aggregation can occur appropriately in FedSOL. Even with diverse clients, the gradient without normalization is simply scaled, thus preserving the effectiveness of aggregation.

\begin{figure}[ht!]
    \centering
    \includegraphics[width=1.0\textwidth]{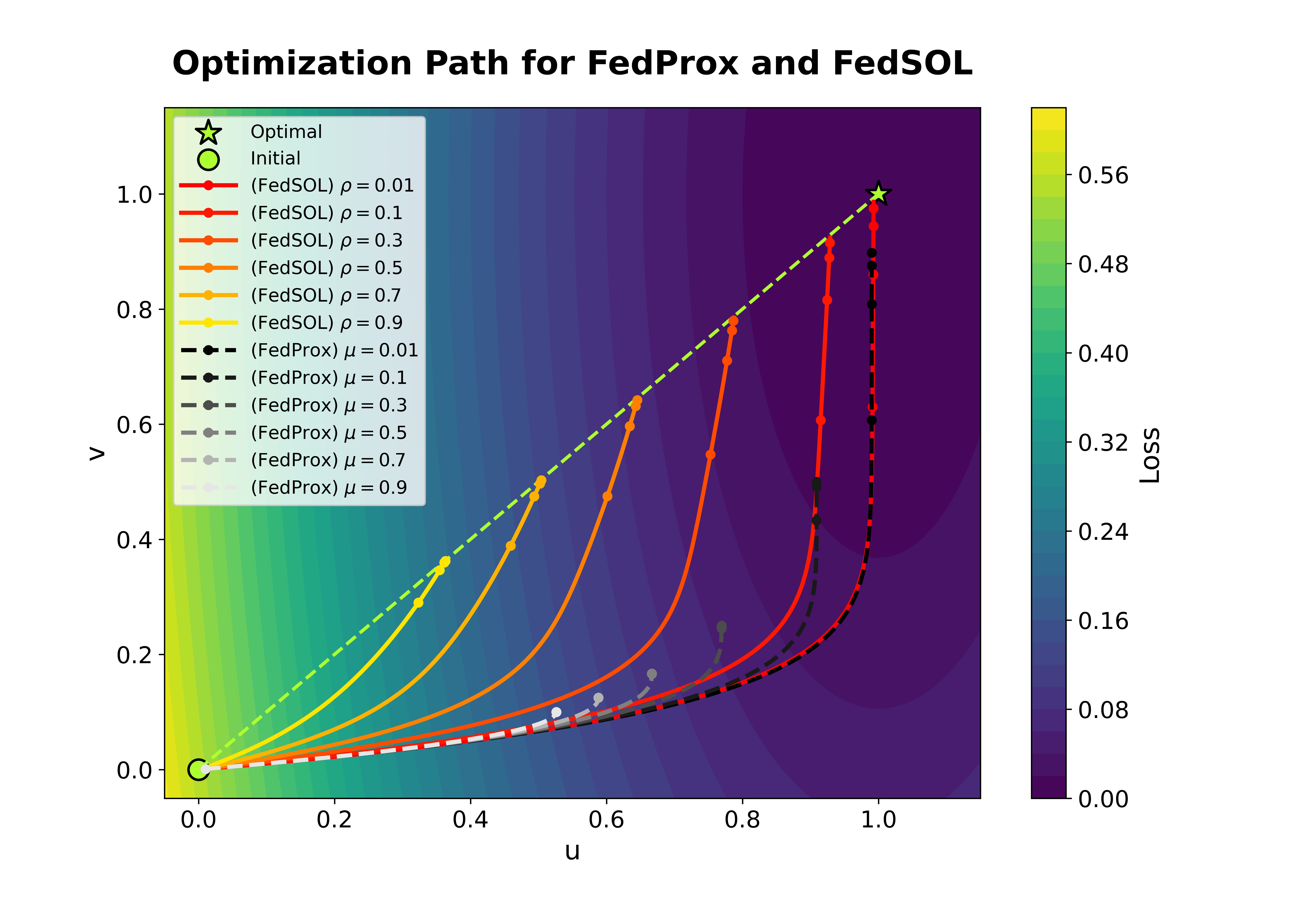}
    \caption{Gradient update paths for FedSOL and FedProx in a toy example on $\delta=0.1$, over 4000 epochs with learning rate $0.01$, and initial weights $(u_l,v_l)=(1,1)$. The illustration highlights the distinct trajectories and convergence points of each algorithm, underlining their differing approaches.}
    \label{fig:fedsol_toy_example}
\end{figure}

\subsection*{Conclusion}
The approach of FedSOL in handling proximal loss demonstrates significant advantages over FedProx within the Federated Learning context. By aligning weight adjustments parallel to the local optimum, FedSOL ensures more efficient and effective aggregation among diverse clients. In contrast, FedProx exhibits a bias towards greater curvature and, as a result, potentially leads to sub-optimal global alignment, as seen with the variable impacts of $\delta$ on different clients. The simulation in \autoref{fig:fedsol_toy_example} corroborates our analysis, showing that the optimized points for FedSOL align with the lime line connecting the initial and optimal points. Meanwhile, FedProx's optimized points are skewed towards the $u$-axis, highlighting the theoretical distinctions between the two algorithms.